\documentclass[article,10pt]{wlscirep}
\usepackage[utf8]{inputenc}
\usepackage[T1]{fontenc}
\usepackage{todonotes}
\usepackage{amsmath}
\usepackage{graphicx}
\usepackage{authblk}
\usepackage{xcolor}
\usepackage{colortbl}
\usepackage{todonotes}
\usepackage{graphicx}
\usepackage[colorlinks=true, allcolors=blue]{hyperref}
\usepackage{multirow}
\usepackage{wrapfig}
\usepackage[table]{xcolor}
\usepackage{url}
\usepackage{subcaption}
\usepackage{devanagari}

\RequirePackage{filecontents}
\usepackage{lineno}
\definecolor{headergray}{gray}{0.80}
\definecolor{rowgray}{gray}{0.80}

\definecolor{maroon}{cmyk}{0,0.87,0.68,0.32}

\title{A Markovian ODE-guided framework for the offline evaluation of reasoning in language models}
\title{Markovian ODE-guided scoring can assess the quality of offline reasoning traces in language models}
\author[1]{Arghodeep Nandi}
\author[1]{Ojasva Saxena}
\author[1,2,3]{Tanmoy Chakraborty\thanks{Corresponding author: tanchak@iitd.ac.in}}

\affil[1]{Department of Electrical Engineering, Indian Institute of Technology Delhi, New Delhi, India}
\affil[2]{Yardi School of Artificial Intelligence, Indian Institute of Technology Delhi, New Delhi, India}
\affil[3]{Indian Institute of Technology Delhi, Abu Dhabi, UAE}

\makeatletter
\renewcommand\@biblabel[1]{}

\makeatother

\begin{abstract}
Reasoning traces produced by generative language models are increasingly used for tasks ranging from mathematical problem solving to automated fact checking. However, existing evaluation methods remain largely mechanical and fail to capture human-centric notions of reasoning quality in a way that generalizes across varied and progressively degraded reasoning. We introduce MarODE, an offline evaluation framework that assigns quality scores to reasoning traces. Its effectiveness is assessed using \emph{human-centric perturbations} and human judgments, which jointly evaluate the fundamental dimensions of an evaluation metric -- \emph{goodness} and \emph{soundness}. The approach is grounded in a Markovian formulation of reasoning progression and an ordinary differential equation based characterization of trace dynamics, enabling efficient evaluation of reasoning quality. In a large-scale evaluation, MarODE outperforms existing baselines by over 250\% under Somers' D correlation. Our results emphasize the value of theory-driven evaluation frameworks as reasoning traces become central to language model-based systems.
\end{abstract}

\begin{document}

\flushbottom
\maketitle

 \section{Introduction}

With the rise of large language models (LLMs), there has been growing interest in leveraging their generative capabilities to produce human-like, multi-step reasoning traces. As such traces become increasingly prevalent, evaluating their quality has emerged as a critical challenge. Recent work shows that this challenge arises because there are few reliable automatic methods to evaluate reasoning chains, and no common framework to systematically compare different reasoning strategies across tasks \citep{hao2024llm}. Reasoning traces, often expressed as chain-of-thought or intermediate rationales, have been shown to substantially improve performance on complex tasks involving arithmetic, symbolic manipulation, multi-hop question answering, and logical inference \citep{wei2022chain,kojima2023largelanguagemodelszeroshot,wang2023selfconsistencyimproveschainthought}. These traces enable models to decompose complex problems into simpler substeps, making them effective across a broad range of domains including mathematics, factual verification, programming, and decision support. Empirical analysis demonstrates that explicitly generating intermediate reasoning steps is a key factor behind these improvements, leading to substantial gains in multi-step reasoning performance even under noisy or imperfect reasoning traces \citep{wang-etal-2023-towards}.
Prior work shows that generated rationales are not always coherent with model's underlying decision process, raising concerns about reliability and interpretability \citep{jacovi-goldberg-2020-towards, yao2023react}. As reasoning traces increasingly mediate interaction between humans and intelligent systems, systematically defining and evaluating their quality remains essential for reliable and effective deployment.

Existing approaches to reasoning evaluation can broadly be divided into two paradigms. \emph{Generation-time} or \emph{online} evaluation assesses reasoning quality during decoding, typically by leveraging intrinsic model signals such as token probabilities, uncertainty estimates, or intermediate verifier feedback to guide or score the generation process \citep{lee-hockenmaier-2025-evaluating}. In contrast, \emph{post-hoc} or \emph{offline} evaluation operates on completed reasoning traces, treating them as textual artefacts and assessing their quality using external criteria such as step-level validity, coherence, factual grounding, and consistency towards an answer. In this work, we focus on post-hoc evaluation of reasoning traces, aiming to characterize and measure the quality of completed reasoning processes independently of the generation mechanism.

Prior studies often struggle to capture a key aspect of reasoning evaluation -- natural language reasoning is inherently human-centric and cannot be reliably assessed using purely mechanical perturbations or metrics designed to detect such perturbations \citep{chrysostomou2021improvingfaithfulnessattentionbasedexplanations}. As language and reasoning styles evolve, evaluation metrics must remain generalizable, rather than being tightly coupled to specific surface level patterns \citep{jacovi-goldberg-2020-towards}. To address this gap, we introduce human-centric perturbations (Section~\ref{subsec:perturbation}) that better reflect realistic variations in reasoning. The perturbation scores serve as ground truth for measuring the correlation of metric scores with deterioration level.  We show that the performance of established benchmarks, such as ROSCOE \citep{golovneva2023roscoe}, degrades sharply under these conditions. Motivated by this observation, we propose \text{MarODE} -- a novel evaluation framework that aims to capture coherence, quality and consistency. Our approach is guided by principled formulations based on Markov chains and consistency probability modeled using ordinary differential equations (ODEs).

Across data distributions, MarODE consistently outperforms existing baselines in sensitivity to perturbations and shows stronger alignment with human judgments on expert-evaluated reasoning traces, supporting its goodness and soundness. Owing to the limited availability of high-quality long-form reasoning traces, we construct traces for perturbation analysis using the claim–verdict–evidence structure of factuality datasets such as LIAR \citep{wang-2017-liar} and PolitiFact \citep{info14120627}. We further assess alignment with human judgments on diverse datasets spanning multiple reasoning domains.

\section{Results}

\paragraph{Formulation of MarODE.}

Evaluating the quality of reasoning traces requires moving beyond final answer correctness or surface-level fluency. Prior work has consistently shown that high-quality reasoning is characterized by coherent step-to-step progression, global directional consistency, and explicit grounding in evidence or premises \citep{dalvi-etal-2021-explaining}. These principles underpin a wide range of reasoning benchmarks and evaluation frameworks, including \textsc{ROSCOE} \citep{golovneva2023roscoe}, EntailmentBank \citep{dalvi-etal-2021-explaining}, ProofWriter \citep{tafjord-etal-2021-proofwriter}, and verifier-based approaches for mathematical and factual reasoning \citep{cobbe2021trainingverifierssolvemath,geva-etal-2021-aristotle}. Guided by these insights, we define reasoning quality as the extent to which intermediate steps form a coherent, non-redundant, logically progressive, grounded in evidence and leading towards an idea.Our formulation of reasoning quality is independent of the final conclusion, emphasizing the direction of idea flow rather than the destination.

{MarODE} directly operationalizes these dimensions within a unified evaluation framework. The first component, Markovian coherence, evaluates local structure by rewarding contiguous logical transitions between steps. This aligns with recent work quantifying semantic alignment and logical errors in chains of reasoning \citep{wang2026chainofthoughtlensevaluatingstructured}. The second component, ODE-guided quality modeling, captures global properties such as redundancy and directional progress and discourages cyclical or repetitive reasoning sequences \citep{chen2025reasoningerasurveylong}. Recent studies have highlighted that language models can produce fluent yet unfaithful explanations that do not reflect their underlying decision processes \citep{turpin2023language}. The third component, evidence alignment, enforces grounding between reasoning steps and relevant evidence, consistent with verifier-based paradigms and process-level validity metrics. The \textit{Methodology} section and Appendices~\ref{subsec:derivation} and \ref{supp:rk4} present a detailed formulation of MarODE.

By integrating these complementary components into a unified score, MarODE provides a comprehensive measure of reasoning trace quality that captures both structural coherence and evidential soundness. This design enables MarODE to serve as a general purpose evaluation metric across factual, logical, and mathematical reasoning, while remaining grounded in established principles of high-quality human reasoning. The complete formulation is referred to below.

\begin{equation}
\boxed{
\begin{aligned}
\mathrm{MarODE}(r)
&= w_c\, \mathbb{E}_{\pi \sim \mathbf{P}}
\left[
\frac{1}{|\pi|-1} \sum_t \mathbb{I}\!\left(|i_{t+1}-i_t|=1\right)
\right] \\[4pt]
&\quad + w_q \left[
\tfrac{1}{2}\!\left(1 - \frac{1}{|\mathcal P|}\!\sum_{(i,j)\in\mathcal P}\rho_{ij}\right)
+ \tfrac{1}{2}\, p_K
\right] \\[4pt]
&\quad + w_e \left[
\frac{1}{K+1}\sum_{i=0}^K
\mathrm{clip}_{[0,1]}
\Big(
0.5 + \lambda_e\, \Pr_i(\mathrm{entail})
- \lambda_c\, \Pr_i(\mathrm{contradict})
\Big)
\right]
\end{aligned}
}
\end{equation}
where $\mathbf{P}$ is the Markov transition matrix over step embeddings, $\pi = (i_0, i_1, \ldots, i_L)$ denote a sampled walk, $\mathcal{P}$ denotes the set of upper-triangular step pairs, $\rho_{ij}$ denotes redundancy penalties, $p_K$ is the ODE-integrated directional belief, and $\Pr_i(\cdot)$ are natural language inference (NLI) probabilities between $R_i$ and evidence.

We present a comprehensive empirical evaluation of MarODE across diverse datasets, models, and evaluation settings. We begin by examining alignment with human-centric perturbations, followed by controlled ablations that disentangle the contributions of individual components. We then assess consistency with expert human judgments on established benchmarks. Finally, we analyze robustness under varying in-context supervision, highlighting stability across shot settings.

\paragraph{Evaluation overview.}

We evaluate {MarODE} against established reasoning quality baselines across datasets, models, and different shot settings. Performance is examined along two dimensions: \emph{goodness}, reflecting sensitivity to controlled degradation in reasoning through human-centric perturbations, and \emph{soundness}, capturing agreement with expert human judgments.
Synthetic reasoning traces (Section~\ref{subsubsec:auto_traces}) are generated using five reasoning optimized LLMs -- \text{DeepSeek-Qwen-14B}, \text{DeepSeek-LLaMA-8B}, \text{DeepSeek-Qwen-7B} \citep{Guo_2025}, \text{Qwen-3B-CoT} \citep{bai2023qwentechnicalreport} and \text{GPT-OSS-20B} \citep{kumar2025gptoss20bcomprehensivedeploymentcentricanalysis}, on 8,600 factual claims drawn from LIAR and PolitiFact, under 1, 2, and 4-shot prompting, yielding 25,800 structured reasoning traces for every model. The prompts used are detailed in Appendix~\ref{supsec:prompt_design}. Evaluation is conducted on both synthetically perturbed reasoning traces and human-annotated benchmarks spanning factual, logical, and mathematical reasoning (Section~\ref{subsubsec:human_traces}). Correlation between metric scores and perturbation levels is quantified using Somers’ $D$. MarODE is compared against ROSCOE, ReCEval, Local and Global Coherence, and an LLM as a Judge, enabling systematic assessment of sensitivity, significance, and alignment across evaluation settings. The details are elaborated in the \textit{Methodology} section (Section~\ref{subsec:experimental_setup}).

\paragraph{Correlation analysis under human-centric perturbations.}

Table~\ref{tab:model_correlations} reports Somers'~$D$ correlations across datasets, prompting settings, and backbone models, measuring alignment with human-centric perturbations. Across LIAR \citep{wang-2017-liar} and PolitiFact \citep{info14120627}, different shot counts, and model families, \text{MarODE} consistently achieves the strongest correlations, outperforming all baselines.
We find that, ROSCOE \citep{golovneva2023roscoe} variants exhibit limited alignment across datasets generated by different model variants. Semantic alignment and similarity (ROSCOE-SA, ROSCOE-SS) yield moderate correlations ($\sim$0.10--0.14), while logical inference (ROSCOE-LI) remains weak and logical consistency (ROSCOE-LC) is near zero or negative, with several non-significant results. Aggregation improves robustness and statistical significance (ROSCOE\_MEAN $\sim$0.06--0.10) but remains markedly below MarODE. Other baselines, including \textsc{LLM-as-a-Judge} \citep{kim2024prometheus2opensource}, \textsc{Local and Global Coherence} \citep{kotonya-toni-2020-explainable-automated}, and \textsc{ReCEval} \citep{prasad2023receval}, consistently show low correlations (typically $<0.05$) and limited sensitivity to increased shot count.
In contrast, MarODE components demonstrate consistently high and statistically significant correlations across datasets, prompting regimes, and models ($\sim$0.24--0.29). Relative to ROSCOE\_MEAN, MarODE yields a $235\%$--$279\%$ increase in correlation strength (Table~\ref{tab:model_correlations}), with gains remaining stable across shot counts and architectural variations. Notably, the relative ordering of metrics is preserved across traces generated by different model families, indicating that MarODE’s improvements are systematic rather than model-specific.

\begin{table}[!h]
\centering
\footnotesize
\rotatebox{0}{


\begin{tabular}{p{2.9cm}*{11}{p{1cm}}}
\rowcolor{headergray}
\toprule
\textbf{Metric} &
\textbf{Qwen-3B-CoT} &
\textbf{DeepSeek-Qwen-7B} &
\textbf{Deepseek-Qwen-14B} &
\textbf{Deepseek-LLaMA-8B} &
\textbf{GPT-OSS-20B} &
\textbf{LIAR (1-shot)} &
\textbf{LIAR (2-shot)} &
\textbf{LIAR (4-shot)} &
\textbf{PolitiFact (1-shot)} &
\textbf{PolitiFact (2-shot)} &
\textbf{PolitiFact (4-shot)} \\
\midrule

ROSCOE-SA
& 0.1187 & 0.1053 & 0.1111 & 0.1117 & 0.1048 & 0.1004 & 0.0983 & 0.1075 & 0.1145 & 0.1152 & \textbf{0.1281} \\

ROSCOE-SS
& 0.1294 & 0.1242 & 0.1301 & 0.1256 & 0.1284 & 0.1167 & 0.1146 & \textbf{0.1386} & 0.1264 & 0.1350 & 0.1331 \\

ROSCOE-LI
& 0.0318 & 0.0163 & 0.0496 & 0.0172 & 0.0146 & 0.0253 & 0.0279 & 0.0286 & 0.0283 & 0.0256 & 0.0200 \\

ROSCOE-LC
& -0.0189 & \underline{-0.0009} & -0.0223 & 0.0228 & -0.0173 & \underline{-0.0073} & \underline{-0.0076} & -0.0184 & \underline{0.0002} & \underline{-0.0047} & \underline{-0.0066} \\
\midrule
ROSCOE\_MEAN
& 0.0819 & 0.0686 & 0.0958 & 0.0689 & 0.0622 & 0.0691 & 0.0728 & 0.0776 & 0.0793 & 0.0785 & 0.0742 \\
\midrule
LLM\_as\_a\_Judge
& 0.0436 & 0.0271 & 0.0421 & 0.0375 & 0.0330 & 0.0367 & 0.0282 & 0.0140 & 0.0516 & 0.0455 & 0.0407 \\
\midrule
Local\_and\_Global
& 0.0458 & 0.0192 & 0.0516 & 0.0321 & 0.0316 & 0.0417 & 0.0397 & 0.0347 & 0.0296 & 0.0379 & 0.0330 \\
\midrule
ReCEval
& 0.0457 & \underline{0.0049} & 0.0311 & \underline{-0.0018} & \underline{0.0072} & \underline{0.0082} & \underline{0.0084} & 0.0211 & 0.0253 & 0.0243 & 0.0177 \\
\midrule
MarODE
& \textbf{0.2937} & \textbf{0.2371} & \textbf{0.2882} & \textbf{0.2921} & \textbf{0.2634} & \textbf{0.2618} & \textbf{0.2636} & \textbf{0.2604} & \textbf{0.2895} & \textbf{0.2840} & \textbf{0.2792} \\
\midrule
MarODE\_COHERENCE($\alpha$)
& \textbf{0.2857} & 0.1747 & \textbf{0.2798} & 0.2826 & 0.2014 & 0.2395 & 0.2357 & 0.2427 & 0.2165 & 0.2284 & 0.2296 \\
MarODE\_QUALITY($\beta$)
& 0.0286 & 0.0639 & 0.0331 & \underline{0.0082} & 0.0467 & 0.0181 & 0.0342 & 0.0358 & 0.0374 & 0.0405 & 0.0410 \\
MarODE\_EVIDENCE($\gamma$)
& 0.2353 & 0.1955 & 0.2253 & 0.2367 & 0.2139 & 0.2075 & 0.2122 & 0.2082 & 0.2398 & 0.2341 & 0.2279 \\
\midrule
MarODE($\alpha\beta$)
& \textbf{0.3272} & \textbf{0.2093} & \textbf{0.3279} & \textbf{0.3118} & \textbf{0.2309} & \textbf{0.2762} & \textbf{0.2743} & \textbf{0.2769} & \textbf{0.2583} & \textbf{0.2652} & \textbf{0.2639} \\
MarODE($\beta\gamma$)
& 0.2403 & 0.1992 & 0.2309 & 0.2392 & 0.2173 & 0.2103 & 0.2164 & 0.2129 & 0.2444 & 0.2379 & 0.2322 \\
MarODE($\alpha\gamma$)
& 0.2857 & \textbf{0.2303} & 0.2791 & \textbf{0.2857} & \textbf{0.2570} & \textbf{0.2550} & \textbf{0.2560} & \textbf{0.2527} & \textbf{0.2799} & \textbf{0.2763} & \textbf{0.2714} \\

\bottomrule
\end{tabular}}
\caption{\textbf{Somers' $D$ correlations measuring the association between human-centric perturbation scores and evaluation metrics across different backbone models and prompting settings.} Columns correspond to reasoning traces generated by a range of LLMs as well as fact-checking benchmarks (\textsc{LIAR} and \textsc{PolitiFact}) under varying shot configurations. Higher values indicate stronger sensitivity to perturbation-induced changes in reasoning quality. For each column, the three highest correlations are highlighted in \textbf{bold}, while underlined values denote correlations that are not statistically significant ($p \geq 0.05$). Results include ROSCOE variants, LLM-based and coherence-oriented baselines, ReCEval, and MarODE with its component-wise and pairwise ablations, enabling a detailed analysis of goodness across models and prompting regimes.}
\label{tab:model_correlations}
\end{table}

\paragraph{Block-wise analysis of MarODE.}

We conduct block-wise ablations of {MarODE} to assess the contribution and interaction of components -- Markovian Coherence ($\alpha$), ODE-guided Quality ($\beta$) and Evidence Alignment ($\gamma$) across datasets and models (Table~\ref{tab:model_correlations}). {\it Markovian Coherence} ($\alpha$) is the dominant factor, achieving high standalone correlations ($\sim$0.24--0.29). Removing other components often improves performance: pairwise combinations excluding evidence grounding, especially ($\alpha\beta$), consistently outperform ($\alpha\gamma$) and attain the strongest pairwise correlations overall ($\sim$0.21--0.33). This suggests that evidence signals can reduce alignment.
\textit{Evidence Alignment} ($\gamma$), shows moderate standalone performance ($\sim$0.20--0.24) but offers limited gains when combined with other blocks. In several cases, adding ($\gamma$) yields marginal or negative changes relative to($\alpha$) based combinations, indicating a secondary or context-dependent role.
\textit{ODE-guided Quality} ($\beta$) has weak standalone alignment (mostly $<0.06$) but acts as a stabilizer when paired with ($\alpha$), improving consistency and correlation strength. However, removing ($\beta$) from the full aggregation can increase correlations, indicating that its benefits are not universal.
Overall, while MarODE remains strong ($\sim$0.24--0.29), ablations show that peak performance often arises from coherence-centered combinations. These results indicate that MarODE's effectiveness is primarily driven by structural coherence, with quality playing a complementary role and evidence providing limited contributions.

\begin{table*}[!t]
\centering
\small
\setlength{\tabcolsep}{4pt}
\renewcommand{\arraystretch}{1.15}

\begin{tabular}{lccccc}
\rowcolor{headergray}
\toprule
\textbf{Metric} & \textbf{ENTAILMENT\_BANK} & \textbf{PROOFWRITER} & \textbf{GSM} & \textbf{STRATQA} \\
\midrule
ROSCOE-SA
& \underline{-0.10337}
& 0.17031
& \underline{-0.04427}
& \underline{0.06290}\\

ROSCOE-SS
& \underline{-0.04991}
& 0.15439
& \underline{-0.03248}
& \underline{0.01250}\\

ROSCOE-LI
& \underline{0.03013}
& \underline{0.05386}
& \underline{-0.02203}
& \underline{0.06036}\\

ROSCOE-LC
& \underline{0.01167}
& \underline{0.09714}
& \underline{-0.05359}
& \underline{0.01563}\\
\midrule
ROSCOE\_MEAN
& \underline{-0.00730}
& 0.21145
& \underline{-0.03248}
& \underline{0.08869}\\
\midrule
LLM\_as\_a\_Judge
& \underline{0.07436}
& 0.00000
& \underline{0.00277}
& \underline{0.00449}\\
\midrule
Local\_and\_Global\_Coherence
& \underline{0.03185}
& \underline{-0.00670}
& \underline{0.01650}
& \underline{0.04894}\\
\midrule
ReCEval
& \underline{-0.02211}
& \underline{-0.02183}
& \underline{\textbf{0.05052}}
& \textbf{0.25611}\\
\midrule
MarODE
& \textbf{0.16981}
& \textbf{0.33896}
& \textbf{0.18578}
& \textbf{0.22563}\\
\midrule
MarODE\_COHERENCE($\alpha$)
& \underline{0.03114}
& 0.21407
& \textbf{0.18086}
& 0.12542\\

MarODE\_QUALITY($\beta$)
& \underline{0.02810}
& \textbf{0.23998}
& \underline{0.02121}
& 0.22133\\

MarODE\_EVIDENCE($\gamma$)
& \textbf{0.16870}
& 0.13692
& -
& \underline{0.07443}\\
\midrule
MarODE($\alpha\beta$)
& \underline{0.02293}
& \textbf{0.33857}
& -
& \textbf{0.27349}\\

MarODE($\beta\gamma$)
& 0.14841
& 0.23823
& -
& 0.19516\\

MarODE($\alpha\gamma$)
& \textbf{0.17732}
& 0.21572
& -
& 0.11604\\

\bottomrule
\end{tabular}
\caption{\textbf{Somers' $D$ correlations measuring alignment between human (manual) judgments and evaluation metrics across four reasoning benchmarks: \textsc{EntailmentBank}, \textsc{ProofWriter}, \textsc{GSM}, and \textsc{StrategyQA}}. Higher values indicate stronger agreement with human assessments. For each dataset, the three highest correlations are highlighted in \textbf{bold}. Underlined values denote correlations that are not statistically significant ($p \geq 0.05$). Results include ROSCOE variants, aggregated and LLM-based baselines, ReCEval, and MarODE along with its component-wise and pairwise ablations, enabling comparison of overall performance as well as the contribution of individual reasoning dimensions.}
\label{tab:humaneval}
\end{table*}

\paragraph{Evaluation against human judgments.}

Table~\ref{tab:humaneval} reports Somers’~$D$ correlations between metric scores and human judgments across four human-evaluated benchmarks: \textsc{EntailmentBank} \citep{dalvi-etal-2021-explaining}, \textsc{ProofWriter} \citep{tafjord-etal-2021-proofwriter}, \textsc{GSM8K} \citep{cobbe2021trainingverifierssolvemath} and \textsc{StrategyQA} \citep{geva-etal-2021-aristotle}. Overall, prior baselines exhibit weak, inconsistent, and often non-significant alignment with human assessments, with several correlations near zero or negative across tasks.
ROSCOE components display highly variable behavior. Although ROSCOE-SA and ROSCOE-SS achieve moderate positive correlations on \textsc{ProofWriter}, their performance degrades on other datasets, including negative correlations on \textsc{EntailmentBank} and \textsc{GSM}, with many results lacking statistical significance. Aggregation (ROSCOE\_MEAN) improves alignment on \textsc{ProofWriter} but remains near zero or negative elsewhere. Other baselines (\textsc{LLM-as-a-Judge}, \textsc{Local and Global Coherence}) and \textsc{ReCEval} similarly show limited alignment, with \textsc{ReCEval} exhibiting strong correlation only on \textsc{StrategyQA}.
In contrast, {MarODE} and its components consistently align with human judgments across benchmarks. While individual dimensions vary by dataset, composite variants improve alignment and statistical significance. The aggregated MarODE achieves the strongest overall alignment, with significant correlations.

\begin{figure}[!t]
    \centering
    \includegraphics[width=\textwidth]{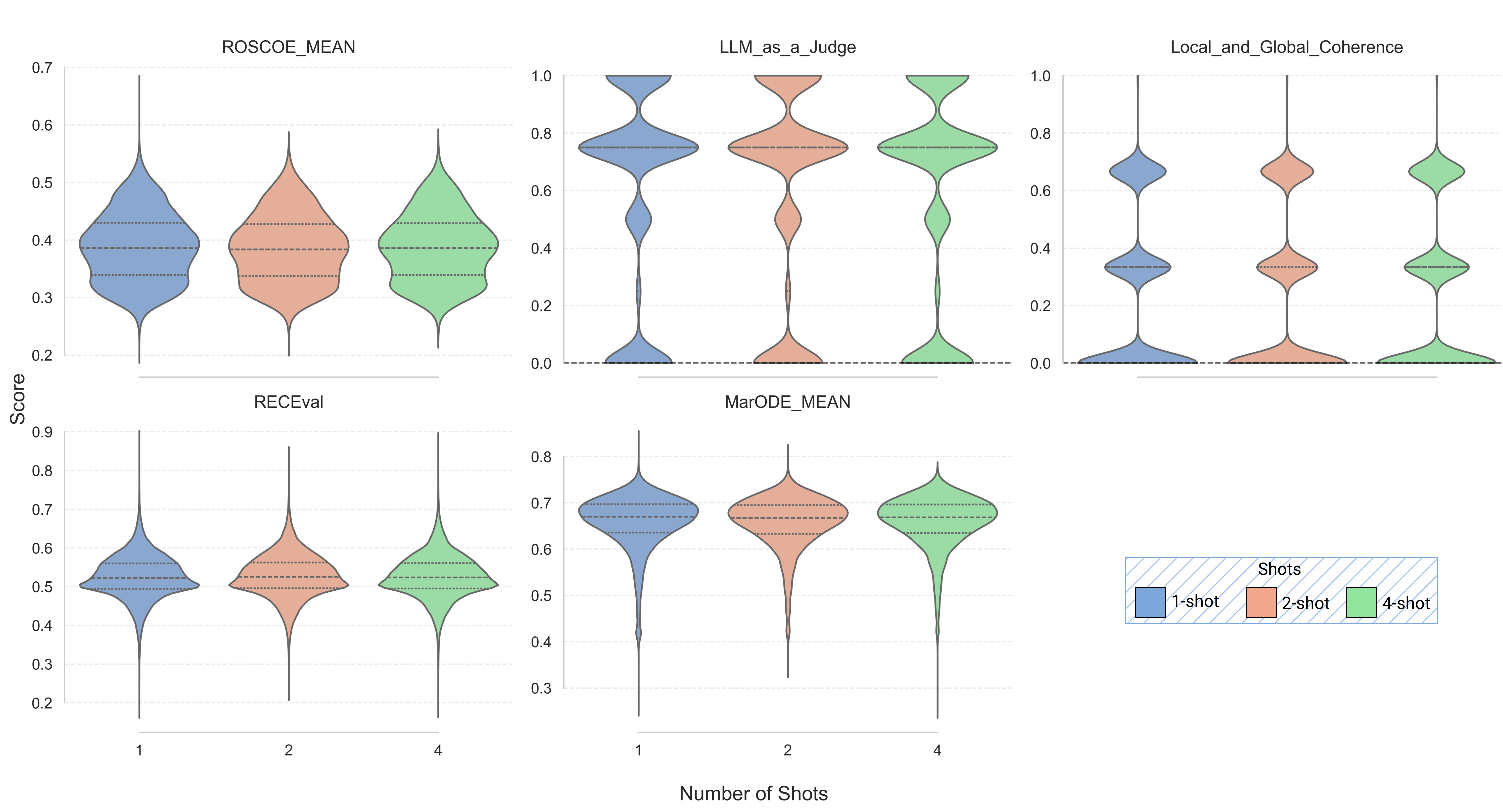}
    \caption{\textbf{Distribution of reasoning trace quality scores across different prompting regimes.} Violin plots illustrate the score distributions for five evaluation metrics -- {ROSCOE}, {LLM-as-a-Judge}, {Local and Global Coherence}, {ReCEval}, and {MarODE}, under three prompting settings (1, 2, and 4-shot) used to generate reasoning traces. For a given baseline, the overall shape of the distribution remains largely consistent across different shot counts, suggesting relative stability of the metric with respect to prompting variation. In contrast, distinct baselines exhibit markedly different distributional shapes, reflecting differences in scale, sensitivity, and scoring behavior across different evaluation approaches.}

    \label{fig:violin}
\end{figure}

\begin{table}[!t]
\centering
\small
\setlength{\tabcolsep}{6pt}
\begin{tabular}{lcccccc}
\rowcolor{headergray}
\hline
\textbf{Baselines} 
& \multicolumn{2}{c}{\textbf{(4-shot -- 1-shot)}} 
& \multicolumn{2}{c}{\textbf{(2-shot -- 1-shot)}} 
& \multicolumn{2}{c}{\textbf{(4-shot -- 2-shot)}} \\
\cline{2-7}
& Median Diff & Significant 
& Median Diff & Significant 
& Median Diff & Significant \\
\hline
ROSCOE-SA          & -0.000769 & TRUE  &  0.000777 & TRUE  & -0.001501 & TRUE  \\
ROSCOE-SS          &  0.001300 & TRUE  & -0.001400 & TRUE  &  0.002700 & TRUE  \\
ROSCOE-LI          & -0.000336 & FALSE & -0.002918 & TRUE  &  0.004092 & TRUE  \\
ROSCOE-LC          & -0.001100 & TRUE  & -0.001100 & TRUE  &  0.000200 & TRUE  \\
\midrule
ROSCOE\_MEAN       & -0.000374 & FALSE & -0.001979 & TRUE  &  0.001659 & TRUE  \\
\midrule
LLM\_as\_a\_Judge  &  0.000000 & FALSE &  0.000000 & TRUE  &  0.000000 & TRUE  \\
\midrule
Local\_and\_Global\_Coherence &  0.000000 & TRUE  &  0.000000 & FALSE &  0.000000 & TRUE  \\
\midrule
ReCEval            &  0.001079 & TRUE  &  0.002021 & TRUE  & -0.001183 & TRUE  \\
\midrule
MarODE           & -0.000479 & TRUE  & -0.001809 & TRUE  &  0.001725 & TRUE  \\
\midrule
MarODE\_COHERENCE($\alpha$) & -0.002344 & TRUE  & -0.000930 & TRUE  & -0.001465 & TRUE  \\
MarODE\_QUALITY($\beta$)   &  0.000188 & TRUE  & -0.000102 & TRUE  &  0.000257 & TRUE  \\
MarODE\_EVIDENCE($\gamma$)  & -0.000658 & FALSE & -0.004585 & TRUE  &  0.003825 & TRUE  \\
\midrule
MarODE($\alpha\beta$)        & -0.000588 & TRUE  & -0.000589 & TRUE  & -0.000115 & FALSE \\
MarODE($\beta\gamma$)        & -0.000095 & FALSE & -0.002524 & TRUE  &  0.002290 & TRUE  \\
MarODE($\alpha\gamma$)       & -0.001165 & TRUE  & -0.002749 & TRUE  &  0.002195 & TRUE  \\
\hline
\end{tabular}
\caption{\textbf{Median score differences and statistical significance ($p < 0.05$) from paired Wilcoxon signed-rank tests across three comparison settings: (4--1), (2--1), and (4--2). }Each setting contrasts reasoning traces generated under different shot conditions, with positive or negative median differences indicating higher or lower metric scores respectively for the former condition in each pair. The \emph{Significant} column denotes whether the observed median difference is statistically reliable. Results are reported for individual ROSCOE variants, aggregated baselines, LLM-based and coherence-focused evaluators, ReCEval, and MarODE along with its component wise and pairwise ablations, enabling a fine-grained analysis of sensitivity to changes in reasoning quality.}

\label{tab:median_significance}
\end{table}

\paragraph{Shot-wise analysis of metric scores.}
\label{sec:shot}

Figure~\ref{fig:violin} analyzes the effect of varying the number of in-context examples (1, 2, and 4-shot prompting) on metric scores for non perturbed reasoning traces. Across all evaluated metrics, changing the shot count does not lead to significant shifts in score distributions. The violins exhibit similar shapes, medians, and inter-quartile ranges, indicating that the overall quality of generated reasoning traces remains stable regardless of the number of shots provided. Wilcoxon signed rank test also failed to conclude any remarkable significant shift in quality of reasoning traces generated using different shots (Table~\ref{tab:median_significance}).
Some baselines, such as \textsc{LLM-as-a-Judge} and \textsc{Local and Global Coherence}, show significant dispersion in central tendencies across different shot settings. Other baselines, such as \textsc{ROSCOE} and \textsc{ReCEval}, while exhibiting lower dispersion, suffer from asymmetric score distributions. In contrast, {MarODE} displays a notably tighter and more symmetric distribution across all shot configurations, reflecting greater stability.

\section{Discussion}
MarODE provides a broader context of evaluating reasoning quality in generated text. Beyond task specific assessment, it provides a general framework for comparing the quality of generated explanations, distinguishing stronger and weaker reasoning patterns, and assessing reliability across models, prompting strategies, and domains. Contemporary work in reasoning evaluation similarly argues for metrics that assess processes rather than only outcomes \citep{Qiu2025, Zhou2025}. By explicitly modeling structural coherence, progression, and evidential support, the approach provides evaluations that are more stable, interpretable, and transferable across reasoning settings, consistent with recent calls for holistic reasoning evaluation beyond task specific metrics \citep{murthy2025metrics}. This relevance becomes increasingly important as LLMs are used to generate multi-step reasoning, where understanding the quality of the reasoning process is as important as the final outcome.

\begin{figure}[htbp]
    \centering
    \includegraphics[width=1\textwidth]{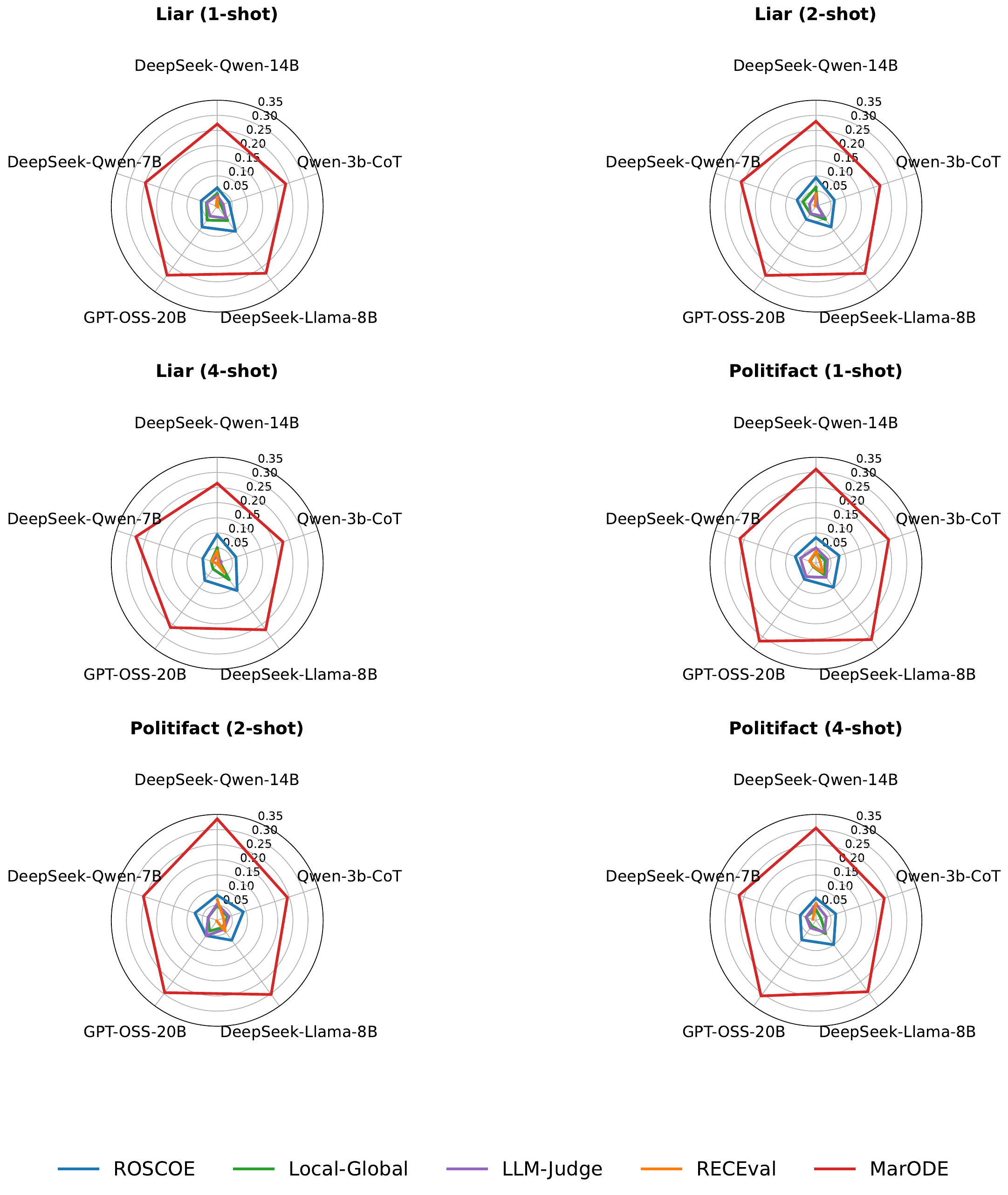}
    \caption{\textbf{Comparative performance of post-hoc reasoning evaluation metrics across datasets and models, visualized as radar charts.} Each subplot corresponds to a specific dataset (LIAR and PolitiFact under 1, 2, and 4-shot prompting), with axes representing the five backbone models. Colored lines denote metric-wise Somers’~$D$ correlations with human-centric perturbations for ROSCOE, Local-Global, LLM-as-a-Judge, ReCEval, and our proposed MarODE. MarODE consistently exhibits the highest correlations across datasets, shot counts, and model families, indicating stronger sensitivity to perturbations that affect reasoning quality.}

    \label{fig:barplot}
\end{figure}

\paragraph{Goodness of MarODE.}
The goodness of an evaluation metric refers to how well it captures meaningful changes in the underlying phenomenon it is designed to measure, especially under realistic variations \citep{yeh-2000-accurate, amidei-etal-2018-evaluation}. For reasoning evaluation, a metric should meaningfully reflect changes in reasoning quality under realistic variations -- responding to differences in logical coherence, evidence use, and inferential consistency, while remaining insensitive to superficial or stylistic changes \citep{lan2025surveyautomaticevaluationmethods, lee-hockenmaier-2025-evaluating}.
Across all experimental settings, {MarODE} consistently demonstrates goodness as an evaluation metric, evidenced by its high and stable correlations with human-centric perturbations (Table~\ref{tab:model_correlations}, Figure~\ref{fig:barplot}). Unlike prior baselines whose correlations remain weak or degrade across datasets, shot counts, and backbone models, MarODE maintains stable performance across diverse conditions.
The magnitude of improvement over existing methods is substantial. Relative to aggregated ROSCOE variants, MarODE achieves a $235\%$--$279\%$ increase in correlation strength, indicating a stronger sensitivity to perturbations that affect reasoning quality rather than surface-level textual properties. Importantly, this improvement is not confined to a specific dataset or model family; the relative ordering of metrics remains preserved across architectures and prompting regimes. This consistency suggests that MarODE possibly captures fundamental properties related to quality of reasoning traces, rather than exploiting dataset specific artefacts.
Taken together, these results indicate that MarODE satisfies a core requirement of post-hoc reasoning evaluation -- it reliably differentiates higher and lower quality reasoning traces under realistic, human centric variations.

\paragraph{Soundness of MarODE.}
Beyond raw performance, a sound reasoning metric should align with human judgments across diverse reasoning domains \citep{chen-etal-2023-rev, deyoung-etal-2020-eraser}. Results on human-evaluated benchmarks (Table~\ref{tab:humaneval}) show that {MarODE} achieves the strongest and most consistent alignment with expert annotations among all evaluated metrics.
Existing baselines, including ROSCOE, LLM-as-a-Judge, and Local–Global Coherence, exhibit highly variable and task-dependent behavior, frequently yielding non-significant or negative correlations. In contrast, MarODE attains statistically significant correlations across four benchmarks spanning symbolic reasoning, mathematical problem solving, multi-hop QA, and strategy-based reasoning. This consistency indicates that MarODE captures reasoning properties that are meaningful to human evaluators rather than overfitting to specific reasoning styles. Together, these findings support MarODE as a general-purpose reasoning evaluation metric whose scores reflect human perceptions of reasoning quality.
Additional evidence for soundness arises from the distributional properties of metric scores. As shown in Figure~\ref{fig:violin}, MarODE exhibits smooth, unimodal, and approximately symmetric distributions, whereas several baselines display skewed, multi-modal, or heavy-tailed patterns. Such irregularities often signal sensitivity to spurious cues, heuristic saturation, or instability in scoring behavior \citep{fomicheva-specia-2019-taking}.
From a statistical perspective, this behavior is characteristic of metrics that implicitly aggregate multiple weakly dependent reasoning factors \citep{longjohn2025statisticaluncertaintyquantificationaggregate}. When evaluation signals integrate diverse underlying properties, score distributions tend to be smoother and more regular, reflecting stable aggregation rather than brittle heuristics \citep{soh2024stepmixturegraderstatistical}. Prior work further shows that metrics with coherent distributional structure align more closely with human judgments than point-wise or threshold based measures \citep{NEURIPS2021_260c2432}, and that well behaved scoring functions yield consistent and interpretable outputs \citep{rei-etal-2023-inside}. In this context, the near-normal shape of MarODE scores suggests balanced integration of reasoning signals rather than dominance by any single criterion.
By contrast, the asymmetric and irregular distributions observed for \textsc{LLM-as-a-Judge}, \textsc{Local and Global Coherence}, and \textsc{ROSCOE} indicate aggregation mechanisms that may be more sensitive to surface cues or specific reasoning styles. The tighter and more symmetric behavior of MarODE supports its interpretation as a stable measure of reasoning quality.

\paragraph{Significance of evidence in reasoning trace evaluation.}

An important insight from our analysis concerns the role of explicit evidence alignment in reasoning evaluation. While evidence grounding is often assumed to be a central component of high-quality reasoning, our results suggest a more nuanced picture.
Evidence alignment ($\gamma$) exhibits moderate standalone performance across perturbation-based evaluations and human judgments. However, its contribution is inconsistent when combined with other components. In several settings, adding evidence alignment to coherence centered formulations yields marginal or even negative effects on correlation strength (Table~\ref{tab:model_correlations}). This pattern indicates that explicit evidence signals can sometimes introduce noise, particularly when evidence relevance or attribution is ambiguous. Empirically, the coherence-quality pairing ($\alpha\beta$) achieves higher correlations than the combination of all three components.
These findings suggest that structural coherence, captured through Markovian transitions, plays a more foundational role in perceived reasoning quality than evidence matching. Evidence acts as a secondary, context dependent signal, strengthening evaluation only when it directly supports intermediate steps. This highlights important implications for metric design, cautioning against over reliance on evidence signals.
An alternative interpretation relates to the perturbations themselves. Reasoning traces are often evaluated in isolation, without access to the full evidence used during generation \citep{prasad2023receval, fang-etal-2024-trace}. As a result, perturbations applied only to reasoning steps may not adequately capture groundedness. Measuring evidence sensitivity may therefore require perturbations that directly target factual consistency and evidence attribution, beyond local, global, or surface-level trace manipulations.

\paragraph{Prompt sensitivity and stability of reasoning quality.}

The shot-wise analysis (Section~\ref{sec:shot}) indicates that varying the number of in-context examples has a limited effect on the quality of generated reasoning traces. As shown by the paired Wilcoxon signed-rank tests in Table~\ref{tab:median_significance}, no consistent or statistically significant differences are observed across shot settings. Across datasets and evaluation metrics, score distributions remain largely similar for 1, 2, and 4-shot prompting, suggesting that increasing demonstrations does not substantially alter reasoning behavior.
This pattern suggests that reasoning quality is driven more by model and task characteristics than by prompt length. While several baseline metrics exhibit variability or skewness across shot settings, MarODE remains comparatively stable, with consistent and symmetric score distributions. This stability indicates that MarODE captures gradual changes in reasoning quality in a controlled manner rather than amplifying prompt-induced fluctuations. Overall, these findings challenge the assumption that increasing in-context examples reliably improves reasoning quality \citep{NEURIPS2024_8cb564df}.

Further evidence of this stability appears in the belief trajectory analysis in Figure~\ref{fig:dircon}. Across diverse reasoning dynamics, including consistently supportive, contradictory, oscillating, and gradual drift scenarios, the ODE-based update underlying MarODE produces smooth and continuous transitions in directional consistency. This behavior aligns with prior work on neural ordinary differential equations, which shows that continuous time formulations yield smooth trajectories and reduced sensitivity to discretization effects \citep{chen2019neuralordinarydifferentialequations}.
In contrast, multiplicative updates often produce abrupt jumps, collapse to extreme values, or exhibit high volatility, particularly in flipping and extreme certainty regimes. Such behavior is consistent with findings from belief and opinion dynamics literature, where multiplicative rules are sensitive to noise, ordering effects, and local inconsistencies, frequently leading to instability or early certainty \citep{10.1109/TSP.2015.2389755}. The stability of the ODE-based formulation is further supported by work on stable neural flows, which demonstrates that continuous dynamical systems with appropriate constraints exhibit bounded, predictable behavior and improved sensitivity to perturbations \citep{massaroli2020stableneuralflows}.

Together, these results indicate that MarODE’s distributional symmetry and stability stem from its continuous modeling of belief updates. By enforcing smooth transitions across reasoning steps, MarODE avoids the abrupt shifts characteristic of discrete multiplicative updates, resulting in consistent evaluation behavior across prompting conditions.

\paragraph{Perturbation sensitivity and metric generalization.}
\label{subsec:roscoe_deterioration}
ROSCOE \citep{golovneva2023roscoe} exhibited perfect correlation across many datasets under its curated perturbations; however, this performance does not generalize when the perturbation set changes. In our experiments, altering the perturbation set leads to severe and inconsistent drops in correlation, indicating that ROSCOE and similar metrics are strongly coupled to the specific mechanical perturbation patterns they are designed to detect \citep{moradi-samwald-2021-evaluating}. As a result, high correlation under a given perturbation set may reflect perturbation specific sensitivity rather than generalized reasoning evaluation.
Effective reasoning metrics should remain largely independent of particular perturbation operators and instead capture general structural properties of reasoning traces \citep{wang-etal-2022-measure}. MarODE is designed to capture this motivation, focusing on coherence, directionality, redundancy, and evidential grounding rather than perturbation specific cues. Although this general formulation yields lower raw correlation scores, MarODE demonstrates higher effectiveness in detecting diverse human centric perturbations and maintains stable behavior under different distributional settings compared to other baselines, making it better suited for generalized reasoning evaluation.

\paragraph{Concluding remarks.}
MarODE achieves the strongest alignment with human-centric perturbations and expert judgments, improving correlation by $235\%$--$279\%$ over prior metrics. We find that coherence acts as the primary driver of reasoning quality, with ODE-guided quality providing additional synergy, while evidence alignment plays a limited role. MarODE also produces smooth and symmetric score distributions compared to baselines, indicating stable evaluation behavior. Overall, these findings show that reasoning quality is strongly determined by the coherent and consistent progression of steps, establishing MarODE as a reliable and general framework for evaluating reasoning traces.

\section{Methodology}
Here, we describe how we construct, perturb, and evaluate reasoning traces, and how these elements come together in MarODE. We begin by outlining the datasets and evaluation setup, followed by an explanation of how reasoning quality is modeled through coherence, progression, and evidence alignment. A key design choice is our use of a continuous, ODE-based formulation to track how reasoning evolves step by step, which remains stable over long reasoning chains where direct probability multiplication can become brittle or overly sensitive. Together, these choices provide a transparent and generalized framework for analyzing reasoning quality across models and settings.

\subsection{Datasets}
\label{subsec:dataset}

Our study evaluates reasoning trace quality across both generated and human evaluated datasets, covering fact verification, logical reasoning, and mathematical reasoning tasks. The dataset construction follows a unified pipeline for reasoning trace generation, extraction, and validation, ensuring comparability across models and prompting conditions.

\subsubsection{Generated reasoning traces}
\label{subsubsec:auto_traces}

We construct a large-scale reasoning trace dataset using two widely adopted fact-checking benchmarks: \textsc{LIAR} \citep{wang-2017-liar} and \textsc{PolitiFact} \citep{info14120627}. Together, these datasets comprise a total of 8,600 factual claims spanning political statements with ground truth veracity labels. Each claim is paired with structured prompts designed to elicit step-by-step reasoning traces.
Reasoning traces are generated using five reasoning optimised large language models: \text{DeepSeek-Qwen-14B}, \text{DeepSeek-LLaMA-8B}, \text{DeepSeek-Qwen-7B} \citep{Guo_2025}, \text{Qwen-3B-CoT} \citep{bai2023qwentechnicalreport} and \text{GPT-OSS-20B} \citep{kumar2025gptoss20bcomprehensivedeploymentcentricanalysis}. For each model, we employ three few-shot prompting configurations of 1-shot, 2-shot, and 4-shot, resulting in three distinct reasoning traces per claim. This yields a total of 25,800 generated reasoning traces across all prompting settings for every model.

Let $\mathcal{C} = \{c_1, \ldots, c_N\}$ denote the set of factual claims, and let $\mathcal{P}_k(c)$ denote the prompt constructed for claim $c$ under a $k$-shot configuration, where $k \in \{1,2,4\}$. For a given language model $m \in \mathcal{M}$, let $p_m(y \mid \mathcal{P}_k(c))$ denote the conditional probability distribution over output sequences induced by $m$ when conditioned on $\mathcal{P}_k(c)$. A generated response is then obtained by sampling
\begin{equation}
y^{(t)}_{m}(c,k) \sim p_m(\cdot \mid \mathcal{P}_k(c)),
\end{equation}
where $t \in \{1,\dots,T\}$ indexes repeated decoding attempts, with a maximum of $T=5$ retries.

\begin{equation}
\mathcal{E}: y \mapsto r,
\end{equation}
Each generated response is post processed using a deterministic extraction operator, which isolates the reasoning span $r$ delimited by explicit reasoning markers. A reasoning trace $r$ is considered valid if and only if it satisfies the structural validity predicate
\begin{equation}
\mathcal{V}(r) = \mathbb{I}\big[r = \{R_0, R_1, \dots, R_K, V_f\},\; K \geq 1 \big],
\end{equation}
where $R_0$ denotes the initial reasoning step, $\{R_1,\dots,R_K\}$ denote intermediate reasoning steps, and $V_f$ denotes the terminal \texttt{Final Verdict}.

If $\mathcal{V}(r)=0$, generation is repeated by incrementing $t \leftarrow t+1$ until either a valid reasoning trace is obtained or $t=T$. For each tuple $(c,k,m)$, the retained reasoning trace is defined as
\begin{equation}
\boxed{
r^{\ast}_{m}(c,k) = \arg\min_{t} \left\{ t \mid \mathcal{V}\!\left(\mathcal{E}\!\left(y^{(t)}_{m}(c,k)\right)\right)=1 \right\}.
}
\end{equation}

The prompts used for the generation of these traces are described in Appendix~\ref{supsec:prompt_design}.

\subsubsection{Human-evaluated reasoning traces}
\label{subsubsec:human_traces}

In addition to model-generated traces, we include a curated set of reasoning traces to serve as high-quality reference data. This subset is drawn from four established reasoning benchmarks: \textsc{EntailmentBank} \citep{dalvi-etal-2021-explaining}, \textsc{ProofWriter} \citep{tafjord-etal-2021-proofwriter}, \textsc{GSM8K} \citep{cobbe2021trainingverifierssolvemath} and \textsc{StrategyQA} \citep{geva-etal-2021-aristotle}. From each dataset, we have 150 instances, resulting in a total of 600 human-evaluated reasoning traces. These datasets span deductive, logical, numerical, commonsense, and scientific reasoning domains, enabling evaluation under heterogeneous reasoning styles.
These datasets cover diverse reasoning phenomena, including multi hop logical entailment, formal proof construction, arithmetic and symbolic reasoning and commonsense question answering. All selected traces follow explicit step-by-step reasoning formats with clearly stated conclusions, making them suitable for fine-grained evaluation of reasoning trace quality under our proposed methodology. 
We evaluate reasoning traces from these datasets using three expert human evaluators. Evaluations are conducted along multiple dimensions, including logical and consistent progression, clarity of expression, absence of redundancy, grounding in the given context, coherence of reasoning flow supporting the final judgment, and overall reasoning quality.

\subsection{Human-centric perturbation}
\label{subsec:perturbation}

To evaluate the sensitivity of reasoning traces under controlled degradation, we apply a set of human-centric perturbations to the generated reasoning traces. Perturbations are designed to selectively disrupt logical structure, semantic coherence, or evidential grounding while preserving surface fluency, thereby enabling fine-grained evaluation of reasoning quality. 

Let $
r = \{R_0, R_1, \ldots, R_K, V_f\}$
denote a reasoning trace consisting of ordered reasoning steps $\{R_i\}_{i=0}^K$ and a terminal verdict $V_f$. Perturbations operate on the step sequence $\{R_i\}_{i=0}^K$, with $V_f$ preserved unless explicitly perturbed.

We define a perturbation function set (details are given in Table~\ref{tab:perturbations}) as
\[
\mathcal{F} = \{f_1, \ldots, f_{14}\},
\]
where each $f_j \in \mathcal{F}$ is a partial operator
\[
f_j : \{R_0,\ldots,R_K\} \rightarrow \{\tilde{R}_0,\ldots,\tilde{R}_K\},
\]
returning a success indicator $\sigma_j \in \{0,1\}$.

To balance perturbation intensity, traces are randomly permuted and assigned a target perturbation count
\[
p \in \{0,\ldots,6\}.
\]
For each trace, up to $p$ distinct perturbations are sampled uniformly without replacement from $\mathcal{F}$ and applied sequentially until
\[
\sum_{j}\sigma_j = p
\quad \text{or} \quad
\mathcal{F} \text{ is exhausted}.
\]

The perturbed trace $\tilde{r}$ is reconstructed with preserved step indices. A perturbation score is defined as
\[
s(p) = 1.0 - 0.1p,\qquad p \in \{0,\ldots,6\},
\]
yielding $s(p) \in [0.4,1.0]$. Each data instance stores $(\tilde{r}, s(p), \{f_j\})$ for downstream evaluation.

\paragraph{Continuous modeling of reasoning degradation.}
In contrast to prior perturbation-based evaluation frameworks such as ROSCOE, which primarily employ binary labeling of perturbed reasoning traces, our methodology models reasoning degradation as a gradual and structured process. Through randomly composed and graded human-centric perturbations targeting local step inconsistencies, global structural disruptions, and surface-level noise, we model gradual degradation in reasoning that reflects partial logical breakdowns, semantic drift, and evidential weakening observed in human reasoning \citep{dasgupta2024languagemodelshumanlikecontent}. The stochastic application of multiple perturbation operators prevents biases from fixed transformation patterns, enabling fine-grained assessment of reasoning robustness while preserving surface fluency. This design supports nuanced evaluation of reasoning quality beyond discrete failure detection.

\begin{figure}[!t]
    \centering
    \includegraphics[width=\textwidth, trim=20 8 8 8, clip]{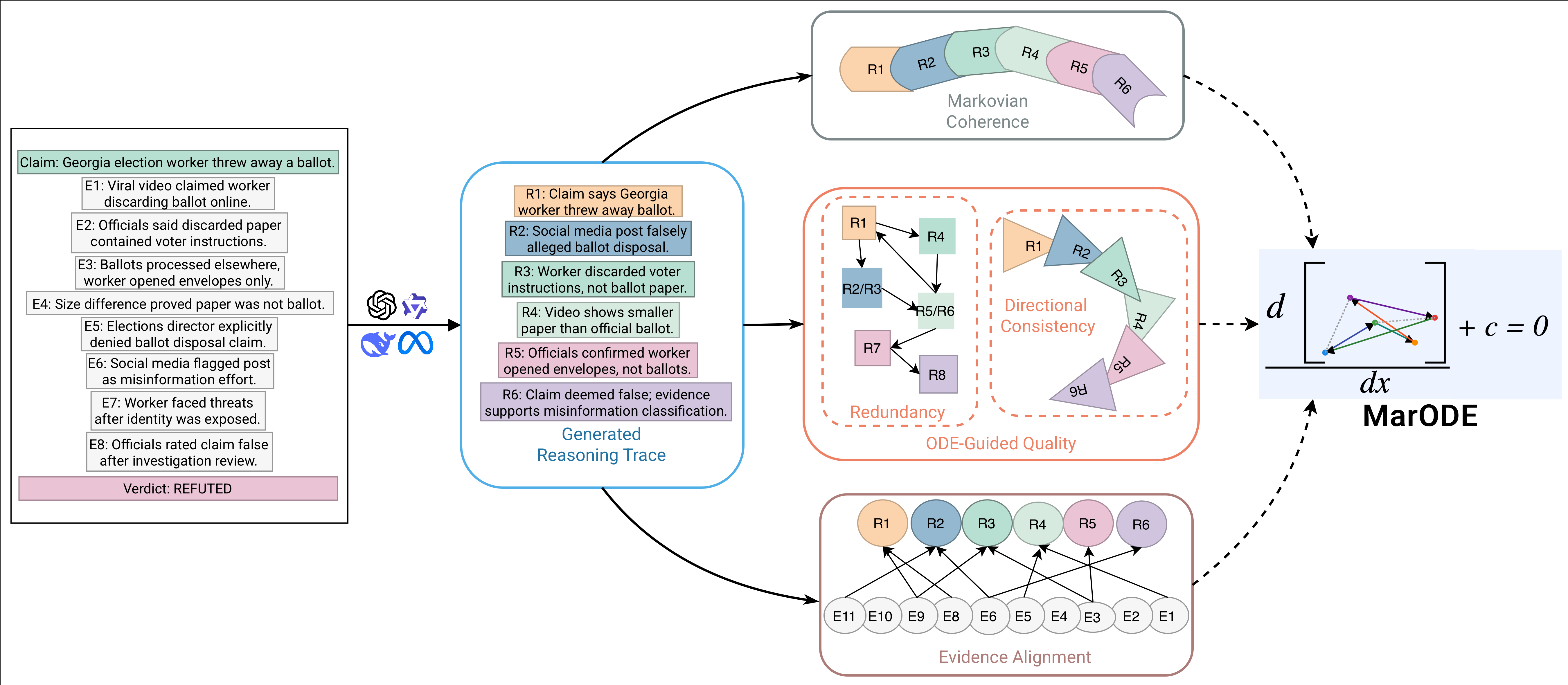}
    \caption{\textbf{An overview of the \textbf{MarODE} framework}, which integrates three complementary components -- \textit{Markovian Coherence}, \textit{Directional Consistency}, and \textit{Evidence Alignment}, to capture a generalist notion of completeness and soundness in reasoning traces. 
    (a) \textit{Markovian Coherence} models the local flow of reasoning, where each step in a high-quality reasoning trace is meaningfully connected to its predecessor while also facilitating progression toward a successor, often with controlled informational overlap. 
    (b) \textit{Directional Consistency} enforces a crisp, one-directional progression of reasoning, ensuring that each step advances toward a conclusion rather than oscillating or regressing. 
    (c) \textit{Redundancy}, Redundant steps that fail to contribute significant new information not only increase the length of the reasoning chain but also compromise its overall quality and its ability to converge on a certain conclusion. 
    (d) \textit{Evidence Alignment} grounds reasoning steps in supporting evidence, mitigating hallucinations that can otherwise lead to confident yet incorrect conclusions.}

    \label{fig:method}
\end{figure}

\subsection{MarODE: Markovian--ODE Reasoning Evaluation}
\label{subsec:marode}

We introduce \textsc{MarODE}, a modular metric for evaluating the quality of reasoning traces. Given a reasoning trace
$r = \{R_0, R_1, \ldots, R_K, V_f\}$,
MarODE computes quality through three components: coherence, quality, and evidence alignment, which are aggregated into a final score. Figure~\ref{fig:method} presents an overview of MarODE.

\subsubsection{Markovian coherence}
\label{subsubsec:marode_coherence}

A good reasoning trace ensures coherent connections between adjacent trace elements. The flow of information should be gradual and continuous, avoiding abrupt or non-justifiable jumps.
Each reasoning step $R_i$ is embedded into a normalized semantic space, yielding embeddings $\{\mathbf{e}_i\}_{i=0}^K$. A transition matrix $\mathbf{P} \in \mathbb{R}^{K \times K}$ is constructed using cosine similarity, 
\begin{equation}
\boxed{
P_{ij} = \frac{\exp\left(\cos(\mathbf{e}_i, \mathbf{e}_j)/\tau\right)}{\sum_{j'} \exp\left(\cos(\mathbf{e}_i, \mathbf{e}_{j'})/\tau\right)},
}
\end{equation}
where $\tau > 0$ denotes a temperature parameter and self-transitions are suppressed. 

This formulation is inspired by prior works that leverages random walks over semantic representations to capture coherence and relatedness, particularly the random-walk–based semantic similarity framework of \citep{10.5555/1708124.1708131} and markov process–based models of document coherence over discourse entities proposed by \citep{petersen2015entropygraphbasedmodelling}. Reasoning coherence is estimated by simulating 256 random walks over the induced Markov chain. Let, $\pi = (i_0, i_1, \ldots, i_L)$ denote a sampled walk. Each walk is scored by rewarding monotonic transitions, where $i_{t+1} = i_t \pm 1$. To obtain a higher quality set of walks, we ensure that the start state lies within the first third of the reasoning steps. The coherence score is defined as
\begin{equation}
\boxed{
C(r) = \mathbb{E}_{\pi \sim \mathbf{P}} \left[ \frac{1}{|\pi|-1} \sum_{t} \mathbb{I}\big(|i_{t+1}-i_t| = 1\big) \right],
}
\end{equation}
approximated via Monte Carlo averaging. This value is recorded as {coherence score}.

\subsubsection{ODE-guided quality modeling}
\label{subsubsec:marode_quality}

Global reasoning quality is modeled by combining redundancy analysis with directional consistency through a continuous time formulation.

\paragraph{Redundancy.}
Each reasoning step $R_i$ is tokenized into a set of alphanumeric tokens $\mathcal{T}(R_i)$. For all unordered step pairs $(i,j)$, a redundancy penalty is computed using symmetric differences,
\[
d_{ij} = \left|\mathcal{T}(R_i) \triangle \mathcal{T}(R_j)\right|.
\]
Penalties are assigned as
\[
\boxed{
\rho_{ij} =
\begin{cases}
1, & d_{ij} = 0, \\
\exp(-d_{ij}/4), & \text{otherwise}.
\end{cases}
}
\]
The redundancy score is then defined as
\begin{equation}
\boxed{
R(r) = 1 - \frac{1}{|\mathcal{P}|} \sum_{(i,j) \in \mathcal{P}} \rho_{ij},
}
\end{equation}
where $\mathcal{P}$ denotes the set of upper-triangular step pairs. This formulation penalizes reasoning traces that repeat the same or very similar words across steps, encouraging each step to add new information. As a result, higher redundancy scores indicate more lexical variety and less repetition throughout the reasoning trace. 

\paragraph{Directional consistency via ODE integration.}
Directional Consistency is modeled by assuming that reasoning trace elements follow a uniform direction toward a specific conclusion. This direction is represented as the evolution of a latent belief variable $p(t) \in (0,1)$, initialized at $p(0)=0.5$. For each reasoning step $R_i$, a Natural Language Inference (NLI) model produces entailment and contradiction probabilities with respect to $R_{i-1}$, yielding a normalized signal:
\begin{equation}
S_i = \mathrm{clip}\!\left(\frac{(e_i - c_i) + 1}{2},\, 0,\, 1\right).
\end{equation}
Here, $e_i$ and $c_i$ denote the entailment and contradiction probabilities, respectively. We employ an ordinary differential equation (ODE) formulation to compute the latent belief variable, rather than using multiplicative probability chains from classical probability theory. Inspired by prior work that formulates latent state trajectories through continuous dynamics, we adapt the continuous time framework of neural ODEs \citep{chen2019neuralordinarydifferentialequations} and recent applications of neural differential equations for latent sequence dynamics \citep{li2025hdndesneuraldifferentialequations} to define directional consistency via ODE integration over reasoning step signals. The rationale and derivation for the following design choice is discussed in  Appendix~\ref{subsec:derivation}.

\[
\boxed{
\frac{dp}{dt} = \rho (S_i - 0.5)p(1-p),
}
\]
Here, $p \in [0,1]$ denotes the evolving belief strength. We model belief evolution through the continuous dynamics which is numerically integrated across reasoning steps using a stable Runge -- Kutta (RK4) scheme (see Appendix~\ref{supp:rk4}).

The integration yields a sequence $\{p_i\}_{i=0}^K$, with $p_{i+1} = \Phi(p_i, S_i)$, where $\Phi$ denotes the numerical RK4 update operator.
The final directional consistency score is defined as
\[
\boxed{
DC(r) = p_K,
}
\]
where $K$ is the length of the reasoning trace.



\paragraph{Quality aggregation.}
The overall quality score is computed as
\begin{equation}
\boxed{
Q(r) = \tfrac{1}{2} R(r) + \tfrac{1}{2} DC(r),
}
\end{equation}

\subsubsection{Evidence alignment}
\label{subsubsec:marode_evidence}

To assess evidential grounding, each reasoning step $R_i$ is aligned with a set of evidence sentences $\mathcal{E}$. For each step, the most similar evidence sentence is selected based on cosine similarity in embedding space. This evidence alignment formulation is inspired by prior work on evidential grounding that aligns reasoning steps with supporting sentences and refines alignment using natural language inference signals, as explored in ERASER \citep{deyoung-etal-2020-eraser} and FEVER style verification models \citep{nie2018combiningfactextractionverification}. If the similarity exceeds a threshold $\theta$, a base alignment score is assigned and refined using NLI probabilities,
\begin{equation}
E_i = 0.5 + \lambda_e\, \Pr_i(\text{entail}) - \lambda_c\, \Pr_i(\text{contradict}),
\end{equation}
clipped to the interval $[0,1]$. The evidence score is defined as
\begin{equation}
\boxed{
E(r) = \frac{1}{K+1} \sum_{i=0}^K E_i,
}
\end{equation}

\subsubsection{Final MarODE aggregation}
\label{subsubsec:marode_final}

The final MarODE score is computed as a convex combination of the three components,
\begin{equation}
\boxed{
\mathrm{MarODE}(r) =
w_c\, C(r) + w_q\, Q(r) + w_e\, E(r)
\label{eq:marode_def}
}
\end{equation}
with default weights $w_c = w_q = w_e = \tfrac{1}{3}$. MarODE provides a unified measure of reasoning coherence, quality, and evidential grounding.

\subsection{Experimental setup}
\label{subsec:experimental_setup}

We evaluate the quality of reasoning traces using a Markovian-ODE based metric and compare it against established baselines. The evaluation relies on synthetic perturbation-based labelling and human judgments as gold standards to assess the goodness and soundness of the proposed metric. Our analysis spans multiple datasets, prompting regimes model families and evaluation metrics.

\subsubsection{Evaluation for soundness and goodness}
We perform experiments under two complementary settings:

\begin{itemize}
    \item \textbf{Perturbation-based labeling (generated diagnostics dataset).} 
    We construct labeled datasets by applying controlled human-centric perturbations (Section \ref{subsec:perturbation}) to generated reasoning traces and assigning graded quality labels according to the degree of perturbation. As the number and severity of perturbations increase, the expected quality of a reasoning trace decreases monotonically \citep{ribeiro-etal-2020-beyond}. This setup enables scalable and controlled evaluation of a metric's sensitivity to systematic degradation in reasoning quality \citep{gardner-etal-2020-evaluating}. We then score the perturbed traces using the proposed metric and competing baselines. We assess alignment between metric scores and perturbation based quality labels using Somers' D to evaluate sensitivity to perturbation induced reasoning degradation.

    \item \textbf{Human-annotated labeling (existing benchmarks).}
    We evaluate alignment with human judgment using expert evaluated reasoning traces drawn from established benchmarks and quantify this alignment using Somers' D correlation. We prepare a set of questions to obtain a holistic human judgment of the quality of each reasoning trace (Figure~\ref{fig:humaneval}). These expert annotations serve as an external gold standard for assessing both the goodness of the proposed metric and its validity in real-world reasoning evaluation settings. This evaluation complements synthetic perturbation based analyses by grounding metric performance in expert human assessments of reasoning quality \citep{chen-etal-2023-rev}.
\end{itemize}

\subsubsection{In-Context supervision and reasoning quality}

To examine the impact of contextual supervision on reasoning quality, we conduct a systematic shot-wise analysis across all datasets, models, and evaluation metrics. Reasoning traces are generated under $1$-shot, $2$-shot, and $4$-shot prompting configurations, and metric scores are computed on the corresponding original traces. Shot-wise trends are quantified via pairwise comparisons between $(4\text{-shot}, 1\text{-shot})$, $(2\text{-shot}, 1\text{-shot})$, and $(4\text{-shot}, 2\text{-shot})$ configurations. We employ the Wilcoxon signed rank test to assess whether observed differences between shot settings are statistically significant \citep{dror-etal-2018-hitchhikers}. This analysis enables us to evaluate whether increasing the number of in-context examples systematically influences the quality of generated reasoning traces \citep{NEURIPS2020_1457c0d6}, and whether such effects are consistent across different evaluation metrics. We further analyze score distributions by plotting violin plots for reasoning traces generated under different shot settings. The shape of each distribution is then used to examine how shot counts influence the quality of the resulting reasoning traces across all baselines.

\subsubsection{Baselines}

We compare {MarODE} against several established baselines that evaluate reasoning traces in an offline setting. Below, we briefly describe the baselines used in our experimental study. Details are provided in the Appendix~\ref{supp:baselines}.
\textbf{ROSCOE} \citep{golovneva2023roscoe} is a collection of interpretable, reference-free metrics designed to assess the quality of step-by-step reasoning, focusing on semantic alignment (ROSCOE-SA), semantic similarity 
(ROSCOE-SS), logical inference (ROSCOE-LI), and language coherence (ROSCOE-LC), using theory-driven error categories and diagnostic datasets. \textbf{ReCEval} \citep{prasad2023receval} evaluates reasoning chains in terms of both correctness and informativeness, leveraging entailment and information-based measures to quantify how individual reasoning steps contribute to the final answer. Prior work \citep{kotonya-toni-2020-explainable-automated} evaluates explanation quality through the notions of \emph{local} and \emph{global coherence}, measuring whether individual reasoning steps are well justified and whether the explanation forms a coherent whole. We also include an \textbf{LLM-as-a-Judge} baseline, in which a large language model is prompted to directly assign a scalar quality score to each reasoning trace using implicitly learned evaluation heuristics.

For entailment-based metrics and related natural language inference components, we use the \texttt{deberta-xlarge-mnli} model across all baselines and MarODE, due to its strong NLI performance. For sentence embeddings, we use the \texttt{all-MiniLM-L6-v2} \citep{Yin2024} model across all baselines and MarODE. The model was chosen for its strong performance across semantic similarity, retrieval, and clustering while maintaining high embedding quality with low-latency, efficient inference. For the LLM-as-a-Judge, we use \texttt{prometheus-7b-v2.0}, chosen for its comparable evaluation quality to larger open models such as GPT-4 \citep{kim2024prometheus2opensource}. For ROSCOE’s linguistic consistency (ROSCOE-LC) components, we use models including \texttt{gpt2-large} and \texttt{tinybertcola} to capture lexical coherence within reasoning chains. All baselines are evaluated on both perturbation-based and human-annotated datasets using Somers’ $D$ correlation.

\subsubsection{Meta-evaluation using Somers' $D$}

To quantify alignment between metric scores and ground truth labels, derived either from controlled perturbations or expert human evaluations, we employ {Somers' $D$} correlation, following prior work such as ROSCOE. Somers' $D$ is a rank based measure of ordinal association that is well suited for meta evaluation settings involving an \emph{ordinal} ground truth variable and a \emph{continuous} metric score. Somers’ $D$ has been widely used in statistics and applied evaluation settings as an asymmetric, rank based association measure designed for ordinal outcomes \citep{Somers1962ANA, Newson2002ParametersB}.

Formally, Somers' $D$ can be expressed as a normalized form of Kendall’s rank correlation coefficient:
\[
\boxed{
D(Y \mid X) = \frac{\tau(X, Y)}{\tau(X, X)},
}
\]
where $\tau(X, Y)$ denotes Kendall’s rank correlation between the ordinal ground truth variable $X$ and the metric score $Y$. As $\tau(X, X)$ quantifies the number of pairs with unequal $X$ values, Somers’ $D$ corresponds to the difference between the number of concordant and discordant pairs, normalized by the number of pairs for which the ground truth labels differ.

\section*{Data and Code Availability}
The source code for reproducing our results is publicly available at \url{https://github.com/ojasiiitd/MarODE}. The generated synthetic datasets are also open-sourced within the same GitHub repository. Additionally, a PyPI package has been created for easy use of \texttt{MarODE}; installation and usage instructions are provided in Appendix~\ref{app:usemarode}.

\section*{Author Contributions}
A.N. and T.C. conceptualized the idea. A.N. and O.S. ran all the experiments. A.N., O.S. and T.C. analyzed the results and prepared the manuscript. T.C. supervised the entire project.

\section*{Competing Interests}
The authors have no competing interests.

\thispagestyle{empty}

\nocite{*}

\bibliography{sample}

\begin{thebibliography}{}

\bibitem[Agarwal et~al., 2024]{NEURIPS2024_8cb564df}
Agarwal, R., Singh, A., Zhang, L., Bohnet, B., Rosias, L., Chan, S., Zhang, B., Anand, A., Abbas, Z., Nova, A., Co-Reyes, J.~D., Chu, E., Behbahani, F., Faust, A., and Larochelle, H. (2024).
\newblock Many-shot in-context learning.
\newblock In Globerson, A., Mackey, L., Belgrave, D., Fan, A., Paquet, U., Tomczak, J., and Zhang, C., editors, {\em Advances in Neural Information Processing Systems}, volume~37, pages 76930--76966. Curran Associates, Inc.

\bibitem[Amidei et~al., 2018]{amidei-etal-2018-evaluation}
Amidei, J., Piwek, P., and Willis, A. (2018).
\newblock Evaluation methodologies in automatic question generation 2013-2018.
\newblock In Krahmer, E., Gatt, A., and Goudbeek, M., editors, {\em Proceedings of the 11th International Conference on Natural Language Generation}, pages 307--317, Tilburg University, The Netherlands. Association for Computational Linguistics.

\bibitem[Bai et~al., 2023]{bai2023qwentechnicalreport}
Bai, J., Bai, S., Chu, Y., Cui, Z., Dang, K., Deng, X., Fan, Y., Ge, W., Han, Y., Huang, F., et~al. (2023).
\newblock Qwen technical report.
\newblock {\em arXiv preprint arXiv:2309.16609}.

\bibitem[Brown et~al., 2020]{NEURIPS2020_1457c0d6}
Brown, T., Mann, B., Ryder, N., Subbiah, M., Kaplan, J.~D., Dhariwal, P., Neelakantan, A., Shyam, P., Sastry, G., Askell, A., Agarwal, S., Herbert-Voss, A., Krueger, G., Henighan, T., Child, R., Ramesh, A., Ziegler, D., Wu, J., Winter, C., Hesse, C., Chen, M., Sigler, E., Litwin, M., Gray, S., Chess, B., Clark, J., Berner, C., McCandlish, S., Radford, A., Sutskever, I., and Amodei, D. (2020).
\newblock Language models are few-shot learners.
\newblock In Larochelle, H., Ranzato, M., Hadsell, R., Balcan, M., and Lin, H., editors, {\em Advances in Neural Information Processing Systems}, volume~33, pages 1877--1901. Curran Associates, Inc.

\bibitem[Chen et~al., 2023]{chen-etal-2023-rev}
Chen, H., Brahman, F., Ren, X., Ji, Y., Choi, Y., and Swayamdipta, S. (2023).
\newblock {REV}: Information-theoretic evaluation of free-text rationales.
\newblock In Rogers, A., Boyd-Graber, J., and Okazaki, N., editors, {\em Proceedings of the 61st Annual Meeting of the Association for Computational Linguistics (Volume 1: Long Papers)}, pages 2007--2030, Toronto, Canada. Association for Computational Linguistics.

\bibitem[Chen et~al., 2025]{chen2025reasoningerasurveylong}
Chen, Q., Qin, L., Liu, J., Peng, D., Guan, J., Wang, P., Hu, M., Zhou, Y., Gao, T., and Che, W. (2025).
\newblock Towards reasoning era: A survey of long chain-of-thought for reasoning large language models.
\newblock {\em arXiv preprint arXiv:2503.09567}.

\bibitem[Chen et~al., 2018]{chen2019neuralordinarydifferentialequations}
Chen, R. T.~Q., Rubanova, Y., Bettencourt, J., and Duvenaud, D.~K. (2018).
\newblock Neural ordinary differential equations.
\newblock In Bengio, S., Wallach, H., Larochelle, H., Grauman, K., Cesa-Bianchi, N., and Garnett, R., editors, {\em Advances in Neural Information Processing Systems}, volume~31. Curran Associates, Inc.

\bibitem[Chrysostomou and Aletras, 2021]{chrysostomou2021improvingfaithfulnessattentionbasedexplanations}
Chrysostomou, G. and Aletras, N. (2021).
\newblock Improving the faithfulness of attention-based explanations with task-specific information for text classification.
\newblock In {\em Proceedings of the 59th Annual Meeting of the Association for Computational Linguistics and the 11th International Joint Conference on Natural Language Processing (Volume 1: Long Papers)}, pages 477--488.

\bibitem[Cobbe et~al., 2021]{cobbe2021trainingverifierssolvemath}
Cobbe, K., Kosaraju, V., Bavarian, M., Chen, M., Jun, H., Kaiser, L., Plappert, M., Tworek, J., Hilton, J., Nakano, R., et~al. (2021).
\newblock Training verifiers to solve math word problems.
\newblock {\em arXiv preprint arXiv:2110.14168}.

\bibitem[Daheim et~al., 2024]{daheim-etal-2024-stepwise}
Daheim, N., Macina, J., Kapur, M., Gurevych, I., and Sachan, M. (2024).
\newblock Stepwise verification and remediation of student reasoning errors with large language model tutors.
\newblock In Al-Onaizan, Y., Bansal, M., and Chen, Y.-N., editors, {\em Proceedings of the 2024 Conference on Empirical Methods in Natural Language Processing}, pages 8386--8411, Miami, Florida, USA. Association for Computational Linguistics.

\bibitem[Dalvi et~al., 2021]{dalvi-etal-2021-explaining}
Dalvi, B., Jansen, P., Tafjord, O., Xie, Z., Smith, H., Pipatanangkura, L., and Clark, P. (2021).
\newblock Explaining answers with entailment trees.
\newblock In Moens, M.-F., Huang, X., Specia, L., and Yih, S. W.-t., editors, {\em Proceedings of the 2021 Conference on Empirical Methods in Natural Language Processing}, pages 7358--7370, Online and Punta Cana, Dominican Republic. Association for Computational Linguistics.

\bibitem[Dasgupta et~al., 2022]{dasgupta2024languagemodelshumanlikecontent}
Dasgupta, I., Lampinen, A.~K., Chan, S.~C., Sheahan, H.~R., Creswell, A., Kumaran, D., McClelland, J.~L., and Hill, F. (2022).
\newblock Language models show human-like content effects on reasoning tasks.
\newblock {\em arXiv preprint arXiv:2207.07051}.

\bibitem[DeYoung et~al., 2020]{deyoung-etal-2020-eraser}
DeYoung, J., Jain, S., Rajani, N.~F., Lehman, E., Xiong, C., Socher, R., and Wallace, B.~C. (2020).
\newblock {ERASER}: {A} benchmark to evaluate rationalized {NLP} models.
\newblock In Jurafsky, D., Chai, J., Schluter, N., and Tetreault, J., editors, {\em Proceedings of the 58th Annual Meeting of the Association for Computational Linguistics}, pages 4443--4458, Online. Association for Computational Linguistics.

\bibitem[Dror et~al., 2018]{dror-etal-2018-hitchhikers}
Dror, R., Baumer, G., Shlomov, S., and Reichart, R. (2018).
\newblock The hitchhiker{'}s guide to testing statistical significance in natural language processing.
\newblock In Gurevych, I. and Miyao, Y., editors, {\em Proceedings of the 56th Annual Meeting of the Association for Computational Linguistics (Volume 1: Long Papers)}, pages 1383--1392, Melbourne, Australia. Association for Computational Linguistics.

\bibitem[Fang et~al., 2024]{fang-etal-2024-trace}
Fang, J., Meng, Z., and MacDonald, C. (2024).
\newblock {TRACE} the evidence: Constructing knowledge-grounded reasoning chains for retrieval-augmented generation.
\newblock In Al-Onaizan, Y., Bansal, M., and Chen, Y.-N., editors, {\em Findings of the Association for Computational Linguistics: EMNLP 2024}, pages 8472--8494, Miami, Florida, USA. Association for Computational Linguistics.

\bibitem[Fomicheva and Specia, 2019]{fomicheva-specia-2019-taking}
Fomicheva, M. and Specia, L. (2019).
\newblock Taking {MT} evaluation metrics to extremes: Beyond correlation with human judgments.
\newblock {\em Computational Linguistics}, 45(3):515--558.

\bibitem[Gardner et~al., 2020]{gardner-etal-2020-evaluating}
Gardner, M., Artzi, Y., Basmov, V., Berant, J., Bogin, B., Chen, S., Dasigi, P., Dua, D., Elazar, Y., Gottumukkala, A., Gupta, N., Hajishirzi, H., Ilharco, G., Khashabi, D., Lin, K., Liu, J., Liu, N.~F., Mulcaire, P., Ning, Q., Singh, S., Smith, N.~A., Subramanian, S., Tsarfaty, R., Wallace, E., Zhang, A., and Zhou, B. (2020).
\newblock Evaluating models' local decision boundaries via contrast sets.
\newblock In Cohn, T., He, Y., and Liu, Y., editors, {\em Findings of the Association for Computational Linguistics: EMNLP 2020}, pages 1307--1323, Online. Association for Computational Linguistics.

\bibitem[Geva et~al., 2021]{geva-etal-2021-aristotle}
Geva, M., Khashabi, D., Segal, E., Khot, T., Roth, D., and Berant, J. (2021).
\newblock Did aristotle use a laptop? a question answering benchmark with implicit reasoning strategies.
\newblock {\em Transactions of the Association for Computational Linguistics}, 9:346--361.

\bibitem[Golovneva et~al., 2023]{golovneva2023roscoe}
Golovneva, O., Chen, M.~P., Poff, S., Corredor, M., Zettlemoyer, L., Fazel-Zarandi, M., and Celikyilmaz, A. (2023).
\newblock {ROSCOE}: A suite of metrics for scoring step-by-step reasoning.
\newblock In {\em The Eleventh International Conference on Learning Representations}.

\bibitem[Guo et~al., 2025]{Guo_2025}
Guo, D., Yang, D., Zhang, H., Song, J., Wang, P., Zhu, Q., Xu, R., Zhang, R., Ma, S., Bi, X., Zhang, X., Yu, X., Wu, Y., Wu, Z.~F., Gou, Z., Shao, Z., Li, Z., Gao, Z., Liu, A., Xue, B., Wang, B., Wu, B., Feng, B., Lu, C., Zhao, C., Deng, C., Ruan, C., Dai, D., Chen, D., Ji, D., Li, E., Lin, F., Dai, F., Luo, F., Hao, G., Chen, G., Li, G., Zhang, H., Xu, H., Ding, H., Gao, H., Qu, H., Li, H., Guo, J., Li, J., Chen, J., Yuan, J., Tu, J., Qiu, J., Li, J., Cai, J.~L., Ni, J., Liang, J., Chen, J., Dong, K., Hu, K., You, K., Gao, K., Guan, K., Huang, K., Yu, K., Wang, L., Zhang, L., Zhao, L., Wang, L., Zhang, L., Xu, L., Xia, L., Zhang, M., Zhang, M., Tang, M., Zhou, M., Li, M., Wang, M., Li, M., Tian, N., Huang, P., Zhang, P., Wang, Q., Chen, Q., Du, Q., Ge, R., Zhang, R., Pan, R., Wang, R., Chen, R.~J., Jin, R.~L., Chen, R., Lu, S., Zhou, S., Chen, S., Ye, S., Wang, S., Yu, S., Zhou, S., Pan, S., Li, S.~S., Zhou, S., Wu, S., Yun, T., Pei, T., Sun, T., Wang, T., Zeng, W., Liu, W., Liang, W., Gao, W., Yu, W.,
  Zhang, W., Xiao, W.~L., An, W., Liu, X., Wang, X., Chen, X., Nie, X., Cheng, X., Liu, X., Xie, X., Liu, X., Yang, X., Li, X., Su, X., Lin, X., Li, X.~Q., Jin, X., Shen, X., Chen, X., Sun, X., Wang, X., Song, X., Zhou, X., Wang, X., Shan, X., Li, Y.~K., Wang, Y.~Q., Wei, Y.~X., Zhang, Y., Xu, Y., Li, Y., Zhao, Y., Sun, Y., Wang, Y., Yu, Y., Zhang, Y., Shi, Y., Xiong, Y., He, Y., Piao, Y., Wang, Y., Tan, Y., Ma, Y., Liu, Y., Guo, Y., Ou, Y., Wang, Y., Gong, Y., Zou, Y., He, Y., Xiong, Y., Luo, Y., You, Y., Liu, Y., Zhou, Y., Zhu, Y.~X., Huang, Y., Li, Y., Zheng, Y., Zhu, Y., Ma, Y., Tang, Y., Zha, Y., Yan, Y., Ren, Z.~Z., Ren, Z., Sha, Z., Fu, Z., Xu, Z., Xie, Z., Zhang, Z., Hao, Z., Ma, Z., Yan, Z., Wu, Z., Gu, Z., Zhu, Z., Liu, Z., Li, Z., Xie, Z., Song, Z., Pan, Z., Huang, Z., Xu, Z., Zhang, Z., and Zhang, Z. (2025).
\newblock Deepseek-r1 incentivizes reasoning in llms through reinforcement learning.
\newblock {\em Nature}, 645(8081):633–638.

\bibitem[Hao et~al., 2024]{hao2024llm}
Hao, S., Gu, Y., Luo, H., Liu, T., Shao, X., Wang, X., Xie, S., Ma, H., Samavedhi, A., Gao, Q., Wang, Z., and Hu, Z. (2024).
\newblock {LLM} reasoners: New evaluation, library, and analysis of step-by-step reasoning with large language models.
\newblock In {\em First Conference on Language Modeling}.

\bibitem[Hasani et~al., 2022]{Hasani2022}
Hasani, R., Lechner, M., Amini, A., Liebenwein, L., Ray, A., Tschaikowski, M., Teschl, G., and Rus, D. (2022).
\newblock Closed-form continuous-time neural networks.
\newblock {\em Nature Machine Intelligence}, 4(11):992--1003.

\bibitem[Jacovi and Goldberg, 2020]{jacovi-goldberg-2020-towards}
Jacovi, A. and Goldberg, Y. (2020).
\newblock Towards faithfully interpretable {NLP} systems: How should we define and evaluate faithfulness?
\newblock In Jurafsky, D., Chai, J., Schluter, N., and Tetreault, J., editors, {\em Proceedings of the 58th Annual Meeting of the Association for Computational Linguistics}, pages 4198--4205, Online. Association for Computational Linguistics.

\bibitem[Jin et~al., 2019]{jin2019pubmedqadatasetbiomedicalresearch}
Jin, Q., Dhingra, B., Liu, Z., Cohen, W., and Lu, X. (2019).
\newblock Pubmedqa: A dataset for biomedical research question answering.
\newblock In {\em Proceedings of the 2019 conference on empirical methods in natural language processing and the 9th international joint conference on natural language processing (EMNLP-IJCNLP)}, pages 2567--2577.

\bibitem[Kidger et~al., 2020]{NEURIPS2020_4a5876b4}
Kidger, P., Morrill, J., Foster, J., and Lyons, T. (2020).
\newblock Neural controlled differential equations for irregular time series.
\newblock In Larochelle, H., Ranzato, M., Hadsell, R., Balcan, M., and Lin, H., editors, {\em Advances in Neural Information Processing Systems}, volume~33, pages 6696--6707. Curran Associates, Inc.

\bibitem[Kim et~al., 2024]{kim2024prometheus2opensource}
Kim, S., Suk, J., Longpre, S., Lin, B.~Y., Shin, J., Welleck, S., Neubig, G., Lee, M., Lee, K., and Seo, M. (2024).
\newblock Prometheus 2: An open source language model specialized in evaluating other language models.
\newblock In {\em Proceedings of the 2024 Conference on Empirical Methods in Natural Language Processing}, pages 4334--4353.

\bibitem[Kojima et~al., 2022]{kojima2023largelanguagemodelszeroshot}
Kojima, T., Gu, S.~S., Reid, M., Matsuo, Y., and Iwasawa, Y. (2022).
\newblock Large language models are zero-shot reasoners.
\newblock {\em Advances in neural information processing systems}, 35:22199--22213.

\bibitem[Kotonya and Toni, 2020]{kotonya-toni-2020-explainable-automated}
Kotonya, N. and Toni, F. (2020).
\newblock Explainable automated fact-checking for public health claims.
\newblock In Webber, B., Cohn, T., He, Y., and Liu, Y., editors, {\em Proceedings of the 2020 Conference on Empirical Methods in Natural Language Processing (EMNLP)}, pages 7740--7754, Online. Association for Computational Linguistics.

\bibitem[Kumar et~al., 2025]{kumar2025gptoss20bcomprehensivedeploymentcentricanalysis}
Kumar, D., Yadav, D., and Patel, Y. (2025).
\newblock Gpt-oss-20b: A comprehensive deployment-centric analysis of openai's open-weight mixture of experts model.
\newblock {\em arXiv preprint arXiv:2508.16700}.

\bibitem[Lan et~al., 2025]{lan2025surveyautomaticevaluationmethods}
Lan, T., Zhou, Y.-H., Ma, Z.-A., Sun, F., Sun, R.-Q., Luo, J., Tu, R.-C., Huang, H., Xu, C., Wu, Z., et~al. (2025).
\newblock A survey of automatic evaluation methods on text, visual and speech generations.
\newblock {\em arXiv preprint arXiv:2506.10019}.

\bibitem[Lee and Hockenmaier, 2025]{lee-hockenmaier-2025-evaluating}
Lee, J. and Hockenmaier, J. (2025).
\newblock Evaluating step-by-step reasoning traces: A survey.
\newblock In Christodoulopoulos, C., Chakraborty, T., Rose, C., and Peng, V., editors, {\em Findings of the Association for Computational Linguistics: EMNLP 2025}, pages 1789--1814, Suzhou, China. Association for Computational Linguistics.

\bibitem[Li et~al., 2025]{li2025hdndesneuraldifferentialequations}
Li, Q., Geng, J., Chen, Z., Zhu, D., Wang, Y., Ma, C., Lyu, C., and Karray, F. (2025).
\newblock Hd-ndes: Neural differential equations for hallucination detection in llms.
\newblock In {\em Proceedings of the 63rd Annual Meeting of the Association for Computational Linguistics (Volume 1: Long Papers)}, pages 6173--6186.

\bibitem[Longjohn et~al., 2025]{longjohn2025statisticaluncertaintyquantificationaggregate}
Longjohn, R., Gopalan, G., and Casleton, E. (2025).
\newblock Statistical uncertainty quantification for aggregate performance metrics in machine learning benchmarks.
\newblock {\em arXiv preprint arXiv:2501.04234}.

\bibitem[Massaroli et~al., 2020]{massaroli2020stableneuralflows}
Massaroli, S., Poli, M., Bin, M., Park, J., Yamashita, A., and Asama, H. (2020).
\newblock Stable neural flows.
\newblock {\em arXiv preprint arXiv:2003.08063}.

\bibitem[Moradi and Samwald, 2021]{moradi-samwald-2021-evaluating}
Moradi, M. and Samwald, M. (2021).
\newblock Evaluating the robustness of neural language models to input perturbations.
\newblock In Moens, M.-F., Huang, X., Specia, L., and Yih, S. W.-t., editors, {\em Proceedings of the 2021 Conference on Empirical Methods in Natural Language Processing}, pages 1558--1570, Online and Punta Cana, Dominican Republic. Association for Computational Linguistics.

\bibitem[Murthy et~al., 2025]{murthy2025metrics}
Murthy, A.~B., Sanneman, L., and Mink, J. (2025).
\newblock Metrics for holistic evaluation of {LLM} reasoning about action, change, and planning.
\newblock In {\em NeurIPS 2025 Workshop on Evaluating the Evolving LLM Lifecycle: Benchmarks, Emergent Abilities, and Scaling}.

\bibitem[Newson, 2002]{Newson2002ParametersB}
Newson, R.~B. (2002).
\newblock Parameters behind “nonparametric” statistics: Kendall's tau, somers’ d and median differences.
\newblock {\em The Stata Journal}, 2:45 -- 64.

\bibitem[Nie et~al., 2019]{nie2018combiningfactextractionverification}
Nie, Y., Chen, H., and Bansal, M. (2019).
\newblock Combining fact extraction and verification with neural semantic matching networks.
\newblock In {\em Proceedings of the AAAI conference on artificial intelligence}, volume~33, pages 6859--6866.

\bibitem[Petersen et~al., 2015]{petersen2015entropygraphbasedmodelling}
Petersen, C., Lioma, C., Simonsen, J.~G., and Larsen, B. (2015).
\newblock Entropy and graph based modelling of document coherence using discourse entities: An application to ir.
\newblock In {\em Proceedings of the 2015 International Conference on The Theory of Information Retrieval}, pages 191--200.

\bibitem[Pillutla et~al., 2021]{NEURIPS2021_260c2432}
Pillutla, K., Swayamdipta, S., Zellers, R., Thickstun, J., Welleck, S., Choi, Y., and Harchaoui, Z. (2021).
\newblock Mauve: Measuring the gap between neural text and human text using divergence frontiers.
\newblock In Ranzato, M., Beygelzimer, A., Dauphin, Y., Liang, P., and Vaughan, J.~W., editors, {\em Advances in Neural Information Processing Systems}, volume~34, pages 4816--4828. Curran Associates, Inc.

\bibitem[Prasad et~al., 2023]{prasad2023receval}
Prasad, A., Saha, S., Zhou, X., and Bansal, M. (2023).
\newblock Re{CE}val: Evaluating reasoning chains via correctness and informativeness.
\newblock In {\em The 2023 Conference on Empirical Methods in Natural Language Processing}.

\bibitem[Põldvere et~al., 2023]{info14120627}
Põldvere, N., Uddin, Z., and Thomas, A. (2023).
\newblock The politifact-oslo corpus: A new dataset for fake news analysis and detection.
\newblock {\em Information}, 14(12).

\bibitem[Qiu et~al., 2025]{Qiu2025}
Qiu, P., Wu, C., Liu, S., Fan, Y., Zhao, W., Chen, Z., Gu, H., Peng, C., Zhang, Y., Wang, Y., and Xie, W. (2025).
\newblock Quantifying the reasoning abilities of llms on clinical cases.
\newblock {\em Nature Communications}, 16(1):9799.

\bibitem[Ramage et~al., 2009]{10.5555/1708124.1708131}
Ramage, D., Rafferty, A.~N., and Manning, C.~D. (2009).
\newblock Random walks for text semantic similarity.
\newblock In {\em Proceedings of the 2009 Workshop on Graph-Based Methods for Natural Language Processing}, TextGraphs-4, page 23–31, USA. Association for Computational Linguistics.

\bibitem[Rei et~al., 2023]{rei-etal-2023-inside}
Rei, R., Guerreiro, N.~M., Treviso, M., Coheur, L., Lavie, A., and Martins, A. (2023).
\newblock The inside story: Towards better understanding of machine translation neural evaluation metrics.
\newblock In Rogers, A., Boyd-Graber, J., and Okazaki, N., editors, {\em Proceedings of the 61st Annual Meeting of the Association for Computational Linguistics (Volume 2: Short Papers)}, pages 1089--1105, Toronto, Canada. Association for Computational Linguistics.

\bibitem[Ribeiro et~al., 2020]{ribeiro-etal-2020-beyond}
Ribeiro, M.~T., Wu, T., Guestrin, C., and Singh, S. (2020).
\newblock Beyond accuracy: Behavioral testing of {NLP} models with {C}heck{L}ist.
\newblock In Jurafsky, D., Chai, J., Schluter, N., and Tetreault, J., editors, {\em Proceedings of the 58th Annual Meeting of the Association for Computational Linguistics}, pages 4902--4912, Online. Association for Computational Linguistics.

\bibitem[Soh and Zhao, 2024]{soh2024stepmixturegraderstatistical}
Soh, Y.~J. and Zhao, J. (2024).
\newblock A step towards mixture of grader: Statistical analysis of existing automatic evaluation metrics.
\newblock {\em arXiv preprint arXiv:2410.10030}.

\bibitem[Somers, 1962]{Somers1962ANA}
Somers, R.~H. (1962).
\newblock A new asymmetric measure of association for ordinal variables.
\newblock {\em American Sociological Review}, 27:799.

\bibitem[Su and Wu, 2015]{10.1109/TSP.2015.2389755}
Su, Q. and Wu, Y.-C. (2015).
\newblock On convergence conditions of gaussian belief propagation.
\newblock {\em Trans. Sig. Proc.}, 63(5):1144–1155.

\bibitem[Tafjord et~al., 2021]{tafjord-etal-2021-proofwriter}
Tafjord, O., Dalvi, B., and Clark, P. (2021).
\newblock {P}roof{W}riter: Generating implications, proofs, and abductive statements over natural language.
\newblock In Zong, C., Xia, F., Li, W., and Navigli, R., editors, {\em Findings of the Association for Computational Linguistics: ACL-IJCNLP 2021}, pages 3621--3634, Online. Association for Computational Linguistics.

\bibitem[Turpin et~al., 2023]{turpin2023language}
Turpin, M., Michael, J., Perez, E., and Bowman, S.~R. (2023).
\newblock Language models don't always say what they think: Unfaithful explanations in chain-of-thought prompting.
\newblock In {\em Thirty-seventh Conference on Neural Information Processing Systems}.

\bibitem[Wang et~al., 2026]{wang2026chainofthoughtlensevaluatingstructured}
Wang, B., Li, Z., Huang, X., Huang, X., and Dong, Y. (2026).
\newblock Chain-of-thought as a lens: Evaluating structured reasoning alignment between human preferences and large language models.
\newblock {\em arXiv preprint arXiv:2511.06168}.

\bibitem[Wang et~al., 2023]{wang-etal-2023-towards}
Wang, B., Min, S., Deng, X., Shen, J., Wu, Y., Zettlemoyer, L., and Sun, H. (2023).
\newblock Towards understanding chain-of-thought prompting: An empirical study of what matters.
\newblock In Rogers, A., Boyd-Graber, J., and Okazaki, N., editors, {\em Proceedings of the 61st Annual Meeting of the Association for Computational Linguistics (Volume 1: Long Papers)}, pages 2717--2739, Toronto, Canada. Association for Computational Linguistics.

\bibitem[Wang, 2017]{wang-2017-liar}
Wang, W.~Y. (2017).
\newblock ``liar, liar pants on fire'': A new benchmark dataset for fake news detection.
\newblock In Barzilay, R. and Kan, M.-Y., editors, {\em Proceedings of the 55th Annual Meeting of the Association for Computational Linguistics (Volume 2: Short Papers)}, pages 422--426, Vancouver, Canada. Association for Computational Linguistics.

\bibitem[Wang et~al., 2022a]{wang-etal-2022-measure}
Wang, X., Wang, H., and Yang, D. (2022a).
\newblock Measure and improve robustness in {NLP} models: A survey.
\newblock In Carpuat, M., de~Marneffe, M.-C., and Meza~Ruiz, I.~V., editors, {\em Proceedings of the 2022 Conference of the North American Chapter of the Association for Computational Linguistics: Human Language Technologies}, pages 4569--4586, Seattle, United States. Association for Computational Linguistics.

\bibitem[Wang et~al., 2022b]{wang2023selfconsistencyimproveschainthought}
Wang, X., Wei, J., Schuurmans, D., Le, Q., Chi, E., Narang, S., Chowdhery, A., and Zhou, D. (2022b).
\newblock Self-consistency improves chain of thought reasoning in language models.
\newblock {\em arXiv preprint arXiv:2203.11171}.

\bibitem[Wei et~al., 2022]{wei2022chain}
Wei, J., Wang, X., Schuurmans, D., Bosma, M., brian ichter, Xia, F., Chi, E.~H., Le, Q.~V., and Zhou, D. (2022).
\newblock Chain of thought prompting elicits reasoning in large language models.
\newblock In Oh, A.~H., Agarwal, A., Belgrave, D., and Cho, K., editors, {\em Advances in Neural Information Processing Systems}.

\bibitem[Yao et~al., 2023]{yao2023react}
Yao, S., Zhao, J., Yu, D., Du, N., Shafran, I., Narasimhan, K.~R., and Cao, Y. (2023).
\newblock React: Synergizing reasoning and acting in language models.
\newblock In {\em The Eleventh International Conference on Learning Representations}.

\bibitem[Yeh, 2000]{yeh-2000-accurate}
Yeh, A. (2000).
\newblock More accurate tests for the statistical significance of result differences.
\newblock In {\em {COLING} 2000 Volume 2: The 18th International Conference on Computational Linguistics}.

\bibitem[Yin and Zhang, 2024]{Yin2024}
Yin, C. and Zhang, Z. (2024).
\newblock A study of sentence similarity based on the all-minilm-l6-v2 model with “same semantics, different structure” after fine tuning.
\newblock In {\em Proceedings of the 2024 2nd International Conference on Image, Algorithms and Artificial Intelligence (ICIAAI 2024)}, pages 677--684. Atlantis Press.

\bibitem[Zhou et~al., 2025]{Zhou2025}
Zhou, S., Xie, W., Li, J., Zhan, Z., Song, M., Yang, H., Espinoza, C., Welton, L., Mai, X., Jin, Y., Xu, Z., Chung, Y.-H., Xing, Y., Tsai, M.-H., Schaffer, E., Shi, Y., Liu, N., Liu, Z., and Zhang, R. (2025).
\newblock Automating expert-level medical reasoning evaluation of large language models.
\newblock {\em npj Digital Medicine}, 9(1):34.

\end{thebibliography}

\clearpage
\appendix
\section*{Appendix}
\label{appendix}

\section{Derivation of Directional Consistency Dynamics}
\label{subsec:derivation}

Let $\mathcal{R}_i$ denote the $i$-th reasoning step in a sequence.  
We define the local entailment and contradiction probabilities between consecutive reasoning steps as
\begin{equation}
P^e_i = \mathbb{P}(\mathcal{R}_i \models \mathcal{R}_{i-1}), \qquad
P^c_i = \mathbb{P}(\mathcal{R}_i \perp \mathcal{R}_{i-1}),
\end{equation}
where $\models$ and $\perp$ denote entailment and contradiction, respectively.

\paragraph{Cumulative entailment and contradiction.}
The cumulative entailment probability up to step $i$ is defined as
\begin{equation}
E_i = \mathbb{P}(\mathcal{R}_i \models \mathcal{R}_{i-1} \models \dots \models \mathcal{R}_0)
= \prod_{k=1}^{i} P^e_k ,
\end{equation}
Similarly, the cumulative contradiction probability is
\begin{equation}
C_i = \mathbb{P}(\mathcal{R}_i \perp \mathcal{R}_{i-1} \perp \dots \perp \mathcal{R}_0)
= \prod_{k=1}^{i} P^c_k .
\end{equation}

\paragraph{Odds formulation.}
We define the local odds for step $\mathcal{R}_i$ as
\begin{equation}
\Lambda(\mathcal{R}_i) = \frac{P^e_i}{P^c_i}.
\end{equation}
The cumulative odds up to step $i$ are then given by
\begin{equation}
O_i = \frac{E_i}{C_i}
= \frac{P^e_i P^e_{i-1} \dots P^e_1}{P^c_i P^c_{i-1} \dots P^c_1}
= \Lambda(\mathcal{R}_i) \, O_{i-1}.
\end{equation}

Taking logarithms yields the additive form
\begin{equation}
\log O_i = \log O_{i-1} + \log \Lambda(\mathcal{R}_i).
\end{equation}
Letting $L_i = \log O_i$, we obtain
\begin{equation}
L_i - L_{i-1} = \log \Lambda(\mathcal{R}_i)
= \log P^e_i - \log P^c_i .
\label{eq:log_odds_increment}
\end{equation}

\paragraph{NLI-based approximation.}
We approximate the log-odds increment using natural language inference (NLI) scores.  
Let
\begin{equation}
S^{\text{raw}}_i = \text{Entail}_{\text{NLI}}(\mathcal{R}_i, \mathcal{R}_{i-1})
- \text{Contradict}_{\text{NLI}}(\mathcal{R}_i, \mathcal{R}_{i-1}),
\end{equation}
and normalize it to the unit interval:
\begin{equation}
S_i = \max\!\left(0, \min\!\left(1, \frac{S^{\text{raw}}_i + 1}{2}\right)\right).
\end{equation}

We then approximate
\begin{equation}
\log \Lambda(\mathcal{R}_i) \approx c \left(S_i - \frac{1}{2}\right),
\end{equation}
where $c$ is a scaling constant. This approximation preserves the qualitative behavior of $\log P^e_i - \log P^c_i$ and captures the relative dominance of entailment over contradiction using NLI signals.

\paragraph{Continuous formulation.}
Interpreting the reasoning index $i$ as a continuous variable, we rewrite the update equation as
\begin{equation}
\frac{\partial L}{\partial i} = r \left(S_i - \frac{1}{2}\right),
\label{eq:log_odds_ode}
\end{equation}
where $r$ is the rate-of-change constant.

\paragraph{Directional Consistency Probability.}
We define the directional consistency (entailment consistency) probability as
\begin{equation}
\pi_i = \sigma(L_i) = \frac{1}{1 + e^{-L_i}}
= \frac{E_i}{E_i + C_i},
\end{equation}
where $\sigma(\cdot)$ denotes the sigmoid function.

Differentiating $\pi_i$ with respect to $i$ gives
\begin{equation}
\frac{\partial \pi_i}{\partial i}
= \frac{d \sigma(L_i)}{d L_i} \cdot \frac{\partial L_i}{\partial i}
= \pi_i (1 - \pi_i) \cdot \frac{\partial L_i}{\partial i}.
\end{equation}

Substituting from Eq.~\eqref{eq:log_odds_ode}, we obtain the governing equation used in our method:
\begin{equation}
\boxed{
\frac{\partial \pi_i}{\partial i}
= \pi_i (1 - \pi_i)\, r \left(S_i - \frac{1}{2}\right)
}
\end{equation}

This equation models how directional consistency evolves across a reasoning sequence, where $\pi_i$ represents the probability that the sequence up to step $i$ maintains a consistent entailment direction.

\begin{figure}[htbp]
    \centering
    \includegraphics[width=0.9\textwidth]{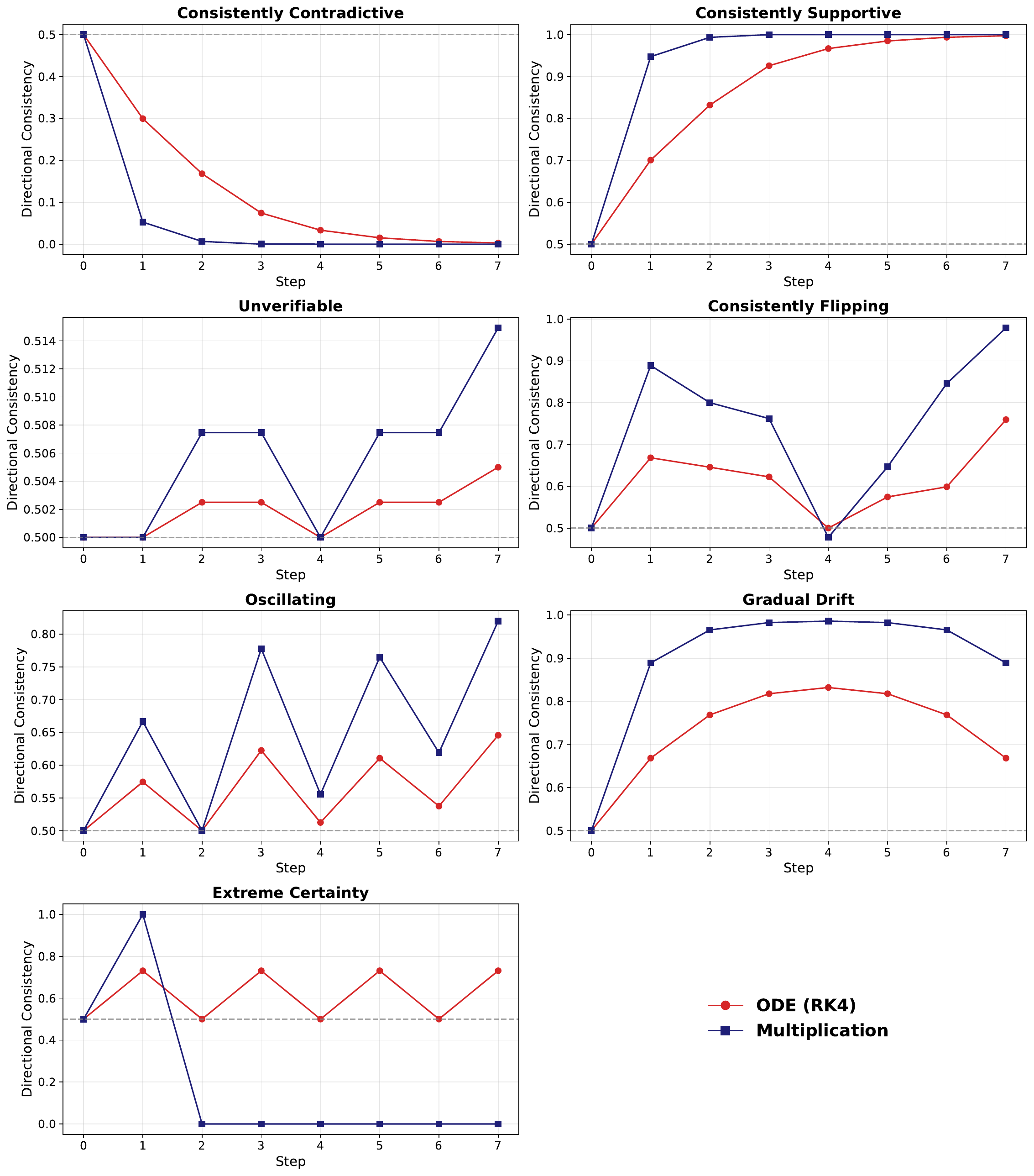}
    \caption{Directional consistencies across seven reasoning scenarios. Each subplot shows the evolution of belief under two update methods: a logistic ODE solved via Runge--Kutta (blue circles) and a multiplicative Bayesian-style update (orange squares). Scenarios include consistently supportive, consistently contradictory, unverifiable, flipping, oscillating, gradual drift, and extreme certainty. The figure highlights differences in stability and volatility across reasoning dynamics.}
    \label{fig:dircon}
\end{figure}

\paragraph{Directional Consistency analysis.} 
To examine how different reasoning dynamics influence directional consistency updating, we simulated seven scenarios of natural language inference (NLI) signals, ranging from consistently supportive to consistently contradictory, oscillating, flipping, gradual drift, and extreme certainty as shown in Figure~\ref{fig:dircon}. For each sequence of entailment, contradiction, and neutral probabilities, we computed directional consistency trajectories using two complementary methods: (i) a logistic ordinary differential equation solved via a fourth order Runge–Kutta scheme, and (ii) a multiplicative Bayesian style update that accumulates entailment and contradiction likelihoods. The resulting plots show how directional consistency evolves step by step under each scenario, highlighting the contrast between \emph{smooth ODE dynamics} and \emph{sharper multiplicative updates}. The ODE-based formulation produces smoother and more steady trajectories, particularly for longer or noisier reasoning chains, while multiplicative updates tend to magnify small local changes and can quickly become extreme. Extensions like neural controlled differential equations further generalize this continuous modeling perspective for dynamic signals \citep{NEURIPS2020_4a5876b4}. Continuous-time modeling has been highlighted in recent machine learning research as a way to capture smoother and more stable temporal dynamics than traditional discrete update methods \citep{Hasani2022}. This comparison highlights why a continuous ODE based approach is better suited for tracking reasoning progression.

\begin{figure}[htbp]
    \centering
    \includegraphics[width=1\textwidth]{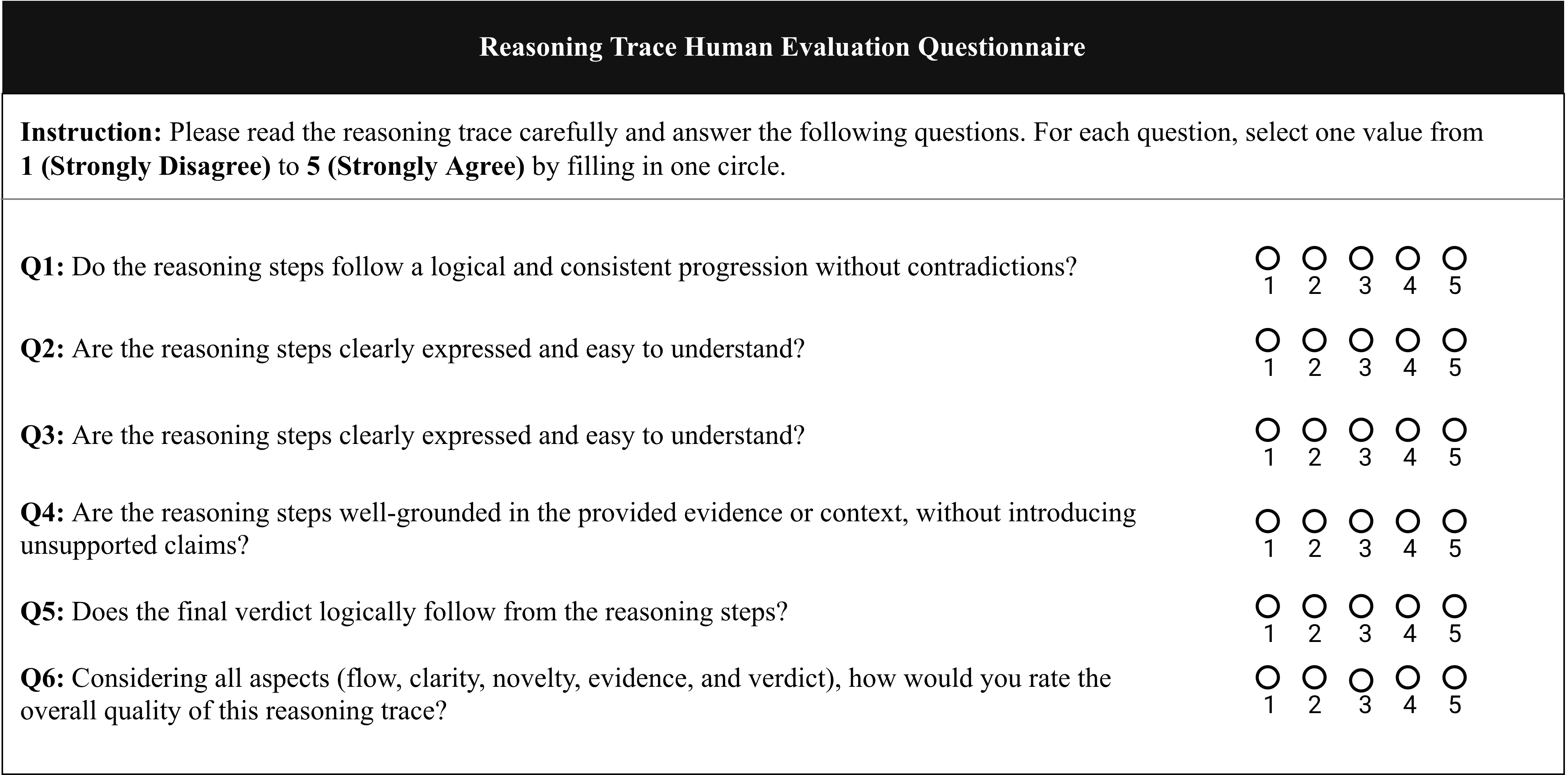}
    \caption{Human evaluation questionnaire used to assess the quality of reasoning traces. Annotators rate each question on a 5-point Likert scale (1 = strongly disagree, 5 = strongly agree).}

    \label{fig:humaneval}
\end{figure}

\begin{table}[h]

\small
\begin{tabular}{@{}llll@{}}
\rowcolor{headergray}
\toprule
\textbf{Perturbation Name} & \textbf{Scope} & \textbf{Description} & \textbf{Failure Mode Targeted} \\
\midrule
Penultimate Ambiguity & Local & Introduces ambiguity in the second-to-last reasoning step & Verdict justification \\
Temporal Confusion & Local & Alters or misorders temporal references & Temporal coherence \\
Unsupported Conclusion & Global & Introduces conclusions not supported by prior steps & Evidence grounding \\
Irrelevant Elaboration & Local & Adds tangential but fluent content & Logical focus \\
Duplication & Local & Repeats reasoning steps verbatim & Redundancy \\
Underspecification & Local & Removes critical explanatory details & Step completeness \\
Antonym Insertion & Local & Injects semantic opposites into key claims & Semantic consistency \\
Definitional Redundancy & Local & Repeats definitions without adding information & Conciseness \\
Modal Logic Confusion & Local & Confuses necessity, possibility, or certainty & Logical modality \\
Ordering & Global & Reorders reasoning steps & Causal structure \\
Quantifier Abuse & Local & Distorts universal or existential claims & Quantitative validity \\
Random Hyphenation & Surface & Inserts spurious hyphenation artifacts & Surface noise \\
Key Concept Swap & Local & Replaces central entities or concepts & Conceptual integrity \\
Final Verdict Insertion & Global & Injects premature verdicts into reasoning & Reasoning closure \\
\hline
\end{tabular}
\caption{Summary of reasoning trace perturbation functions used in this study. Each perturbation targets a specific failure mode in multi-step reasoning, spanning local step level inconsistencies, global structural disruptions, and surface level noise. Perturbations are applied sequentially without replacement to each reasoning trace, with a maximum of six successfully applied perturbations per trace, ensuring controlled yet diverse degradations of reasoning quality.}
\label{tab:perturbations}
\end{table}

\section{Numerical Integration of Directional Consistency Dynamics}
\label{supp:rk4}
To stably integrate the directional consistency ODE

\[
\frac{dp}{dt} = \rho (S_i - 0.5)p(1-p),
\]

We employ a classical fourth-order Runge--Kutta (RK4) method, which provides a balance between numerical accuracy and computational efficiency for nonlinear dynamics.

Given the current belief state $p_i$ and step signal $S_i$, the intermediate slopes are computed as:

\begin{align*}
k_1 &= f(p_i, S_i), \\
k_2 &= f\!\left(p_i + \tfrac{\Delta t}{2}k_1, S_i\right), \\
k_3 &= f\!\left(p_i + \tfrac{\Delta t}{2}k_2, S_i\right), \\
k_4 &= f\!\left(p_i + \Delta t\, k_3, S_i\right),
\end{align*}

where
\[
f(p, S_i) = \rho (S_i - 0.5)p(1-p).
\]

The belief state is then updated as

\[
p_{i+1} = p_i + \frac{\Delta t}{6}\left(k_1 + 2k_2 + 2k_3 + k_4\right),
\]

followed by clipping to the unit interval. This procedure is repeated across all reasoning steps, producing the final directional consistency score $DC(r)=p_K$.

RK4 is widely used for integrating nonlinear dynamical systems due to its fourth-order accuracy and stability properties, making it suitable for modeling smooth belief trajectories across discrete reasoning steps.

\section{Baseline Methodologies}
\label{supp:baselines}

\paragraph{ROSCOE (Reasoning and Stepwise Coherence Evaluation).}
ROSCOE \citep{golovneva2023roscoe} is a modular framework for diagnosing multiple failure modes in chain-of-thought reasoning without requiring gold references. It decomposes reasoning quality into interpretable components capturing semantic alignment, logical validity, and linguistic coherence. ROSCOE-SA and ROSCOE-SS compute embedding-based semantic similarity between adjacent steps and between steps and conclusions to assess topical consistency and semantic progression. ROSCOE-LI leverages natural language inference models to quantify entailment and contradiction relations across reasoning steps, enabling detection of logical breaks or unsupported transitions. ROSCOE-LC evaluates surface-level fluency and syntactic coherence using pretrained language acceptability models. These component scores can be aggregated to yield a holistic quality measure, while maintaining diagnostic transparency across distinct reasoning error categories.

\paragraph{RecEval (Reasoning Chain Evaluation).}
RecEval \citep{prasad2023receval} evaluates reasoning traces by jointly modeling correctness and informativeness of intermediate steps. Logical correctness is estimated using entailment-based verification between reasoning steps and final conclusions, ensuring that intermediate claims are supported. Informativeness quantifies the degree to which each step contributes novel, non-redundant information toward solving the task. By integrating these dimensions, RecEval penalizes reasoning chains that are either logically flawed or unnecessarily repetitive, framing reasoning quality as a contribution-based process rather than purely semantic similarity.

\paragraph{Local and Global Coherence Metrics.}
Local–Global Coherence \citep{kotonya-toni-2020-explainable-automated} approaches assess explanation quality by separating step-wise justification from overall narrative consistency. Local coherence evaluates whether individual reasoning steps are supported by preceding context, typically through semantic similarity or entailment relations between adjacent steps. Global coherence measures whether the complete reasoning trace forms a logically consistent and goal-directed explanation supporting the final conclusion. This decomposition enables detection of both localized reasoning gaps and holistic inconsistencies, though it generally relies on discrete pairwise comparisons without modeling cumulative reasoning dynamics.

\paragraph{LLM-as-a-Judge.}
\label{para:llm_judge}

The \textit{LLM-as-a-Judge} paradigm formulates reasoning trace evaluation as a direct qualitative scoring task performed by a large language model. Given a generated reasoning trace, the evaluator model is prompted to assign a scalar quality score. Rather than relying on explicit symbolic metrics, this approach leverages the implicit reasoning and evaluative capabilities of modern LLMs to approximate human judgment at scale.

\begin{figure}[t]
    \centering
    \includegraphics[width=0.5\linewidth]{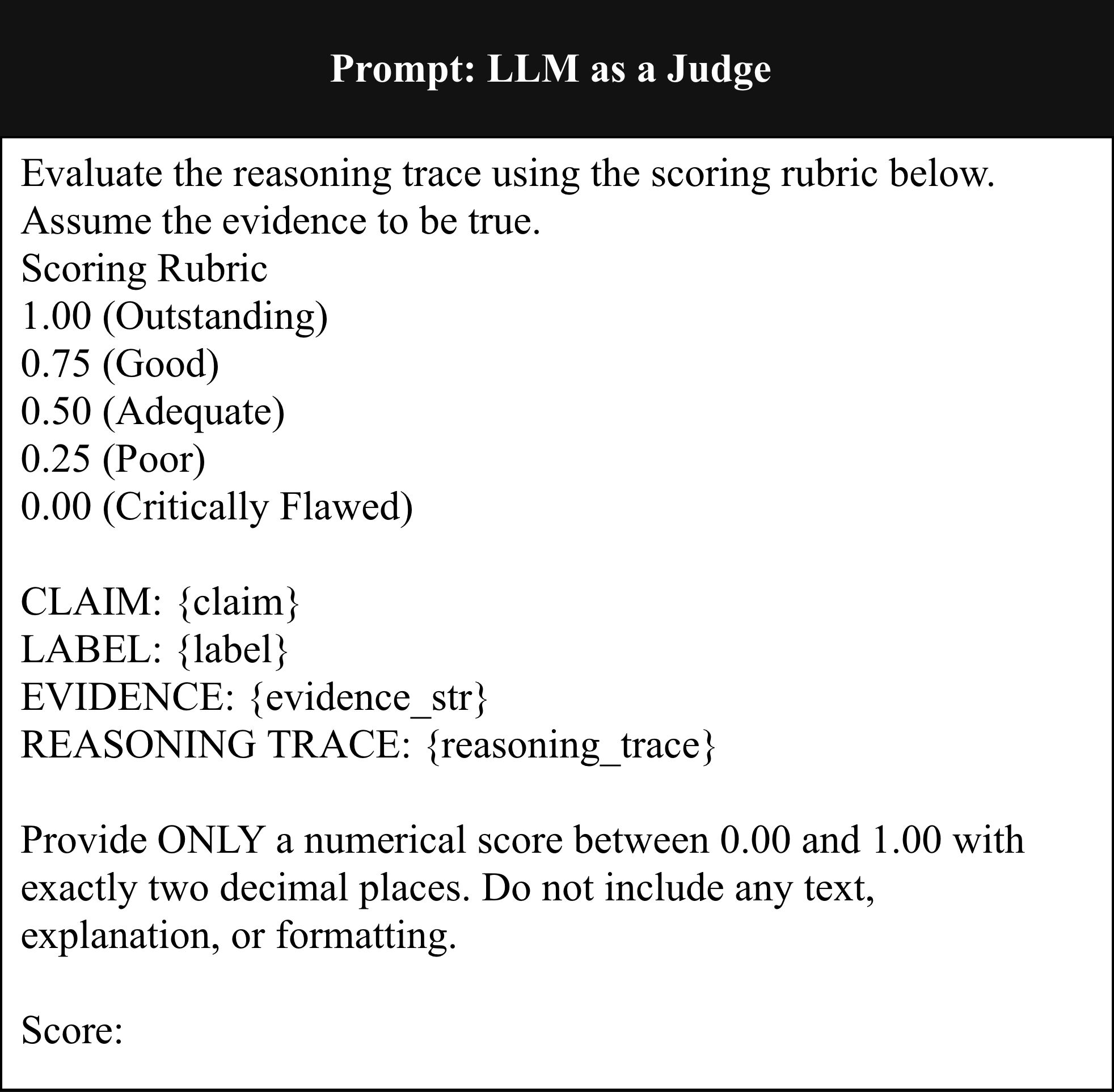}
    \caption{Prompt template used in the LLM-as-a-Judge evaluation paradigm. The evaluator model is instructed to assess the quality of a given reasoning trace and output a single scalar score on a continuous scale, without producing explanatory text. This formulation enables scalable qualitative assessment but relies on the model's implicit evaluation heuristics.}
    \label{fig:llm_judge_prompt}
\end{figure}

As illustrated in Figure~\ref{fig:llm_judge_prompt}, the evaluation prompt constrains outputs to a single continuous numeric score, promoting consistency and ease of aggregation across large trace collections. While effective in practice, this paradigm remains sensitive to prompt phrasing, evaluator model biases, and domain shifts, and offers limited transparency due to reliance on latent evaluation heuristics rather than explicit reasoning criteria.

\section{Prompt Design for Reasoning Trace Generation}
\label{supsec:prompt_design}

\begin{figure}[t]
    \centering
    \includegraphics[width=\linewidth]{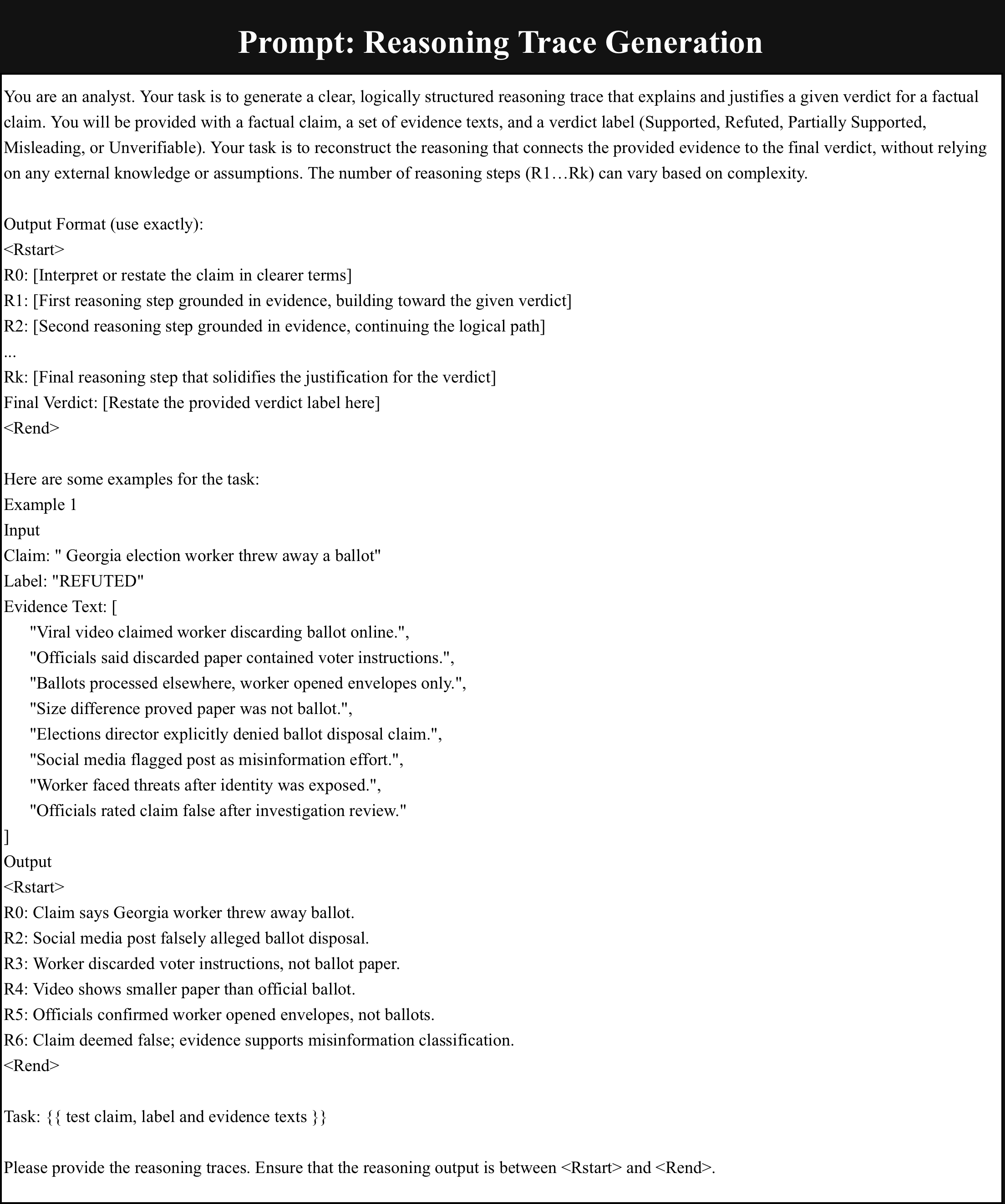}
    \caption{Standardized prompt template used for reasoning trace generation. The prompt provides the factual claim, associated evidence texts, and ground truth verdict label, and constrains model outputs to a structured multi step reasoning format delimited by explicit markers. Few-shot exemplars guide evidence grounding and logical progression toward the final verdict, enabling deterministic extraction and structural validation of generated reasoning traces.}
    \label{fig:trace_prompt}
\end{figure}

To generate structured, evidence-grounded reasoning traces, we employ a standardized prompt template that explicitly constrains both the logical flow and output format of model responses (Figure~\ref{fig:trace_prompt}). The prompt frames the model as an analyst tasked with reconstructing a step-by-step justification linking a factual claim and its associated evidence texts to a provided ground-truth verdict label, without relying on external knowledge or assumptions.

Each prompt instance consists of three components: (i) a factual claim, (ii) a set of evidence snippets retrieved from the fact-checking benchmark, and (iii) the corresponding veracity label. The model is instructed to produce a reasoning trace beginning with an initial claim restatement ($R_0$), followed by a variable-length sequence of evidence-grounded inference steps ($R_1$ through $R_K$), and concluding with a terminal \texttt{Final Verdict} explicitly restating the given label.

To ensure deterministic extraction and structural validity, reasoning spans are delimited using explicit markers \texttt{<Rstart>} and \texttt{<Rend>}. Few-shot exemplars embedded within the prompt illustrate progressive evidence attribution, contradiction resolution, and logical progression toward the verdict. The number of intermediate steps is left unconstrained, allowing trace length to adapt to claim complexity.

\section{Installation of MarODE}
\label{app:usemarode}

\begin{enumerate}
    \item Install the package:
    \begin{verbatim}
    pip install marode
    \end{verbatim}

    \item Import and initialize the evaluator:
    \begin{verbatim}
    from marode.evaluator import MarODEEvaluator, EvaluatorConfig, get_device

    config = EvaluatorConfig()
    device = get_device(-1)  # -1 for CPU, 0 for GPU
    evaluator = MarODEEvaluator(config, device)
    \end{verbatim}

    \item Prepare an entry with reasoning trace and evidence:
    \begin{verbatim}
    entry = {
        "id": "example_1",
        "reasoning_trace": "R0: ...\nR1: ...\nR2: ...",
        "evidence_text": ["evidence 1", "evidence 2"]
    }
    \end{verbatim}

    \item Score the entry and print only the scores:
    \begin{verbatim}
    scored_entry = evaluator.score_entry(entry)
    scores = scored_entry["ourmetric"]
    for k, v in scores.items():
        print(f"{k}: {v:.4f}")
    \end{verbatim}
\end{enumerate}
	
\end{document}